\documentclass{article}

\PassOptionsToPackage{numbers, compress}{natbib}

 \usepackage[final]{neurips_2025}
 \makeatletter
\providecommand{\@trackname}{Epistemic Intelligence in 
Machine Learning}
\makeatother

\usepackage{tikz}
\usepackage{multirow}
\usepackage{pgfplots}
\usepackage{xcolor}
\usepackage{tikz}
\usepackage{pgfplots}

\usepackage{pgfplots}

\pgfplotsset{compat=1.18}

\usepackage{xcolor}
\usetikzlibrary{patterns}

\definecolor{ukrainian}{HTML}{0057B7} 
\definecolor{russian}{HTML}{D62718}   

\usepackage{natbib}
\usepackage[utf8]{inputenc} 
\usepackage[T1]{fontenc}    
\usepackage{hyperref}       
\usepackage{url}            
\usepackage{booktabs}       
\usepackage{amsfonts}       
\usepackage{nicefrac}       
\usepackage{microtype}      
\usepackage{xcolor}         
\usepackage{amsmath}
\usepackage{graphicx}
\usepackage{pgfplots}
\usepackage{pgfplotstable}
\usepackage{caption}
\usepackage{subcaption}
\usepackage{graphicx}
\usepackage{tikz}
\usepackage{pgfplots}
\usetikzlibrary{positioning,patterns,pgfplots.groupplots}
\pgfplotsset{compat=1.18}

\definecolor{bar1}{RGB}{33,113,181}  
\definecolor{bar2}{RGB}{107,174,214} 
\definecolor{bar3}{RGB}{189,215,231} 
\definecolor{bar4}{RGB}{253,174,107} 
\definecolor{bar5}{RGB}{203,24,29}   

\title{Memorization or Reasoning?\\ Measuring epistemic stance Transfer in LLMs via Targeted Unlearning}
\title{Measuring epistemic stance Transfer in LLMs via Targeted Unlearning: A Controlled Test of Reasoning under Epistemic Uncertainty}
\title{Do LLMs Learn to Reason or Just Memorize?\\Testing epistemic stance Transfer via Knowledge Deletion}
\title{Do LLMs Learn to Reason or Just Memorize?\\Testing Epistemic Stance  Transfer via\\Knowledge Deletion}
\title{LLMs Learn How Communities Reason, Not Just What They Say: Testing Epistemic Stance Transfer via Knowledge Deletion}
\title{Stable Response Policies Under Uncertainty:\\Evidence of Epistemic Stance Transfer in LLMs}
\title{Measuring Epistemic Stance Transfer: A Methodology for Community-Aligned Behavior in LLMs}
\title{Community-Aligned Behavior Under Uncertainty:\\Evidence of Epistemic Stance Transfer in LLMs}

\workshoptitle{Epistemic Intelligence in 
Machine Learning}

\author{%
  Patrick Gerard \\
  University of Southern California\\
  Los Angeles, CA 90292 \\
  \texttt{pgerard@isi.edu} \\
  \And
  Aiden Chang \\
  Independent Researcher\\
  \texttt{aidenchang@gmail.com} \\
  \newline
  \And
  Svitlana Volkova \\
  Aptima, Inc.\\
  Woburn, MA 01801 \\
  \texttt{svolkova@aptima.com} \\
}

\begin{document}

\maketitle

\begin{abstract}

When large language models (LLMs) are aligned to a specific online community, do they exhibit generalizable behavioral patterns that mirror that community's attitudes and responses to new uncertainty, or are they simply recalling patterns from training data? We introduce a framework to test \emph{epistemic stance transfer}: targeted deletion of event knowledge, validated with multiple probes, followed by evaluation of whether models still reproduce the community’s organic response patterns under ignorance. Using Russian–Ukrainian military discourse and U.S. partisan Twitter data, we find that even after aggressive fact removal, aligned LLMs maintain stable, community-specific behavioral patterns for handling uncertainty. These results provide  evidence that alignment encodes structured, generalizable behaviors beyond surface mimicry. Our framework offers a systematic way to detect behavioral biases that persist under ignorance, advancing efforts toward safer and more transparent LLM deployments.

\end{abstract}

\section{Introduction}

Large language models (LLMs) now generate human-like text across domains, giving researchers powerful tools to model specific perspectives and community worldviews~\cite{li2023steerability, simmons2023large, rossetti2024social}. Through persona-based prompting or fine-tuning on community data, these models can reproduce discourse patterns~\cite{shi2024saferinstructaligninglanguagemodels, santurkar2023whose, jiang2022communitylm} and are increasingly used in social science, policy analysis, and simulation research~\cite{he2024community, suh2025language}.

However, a central question remains: when aligned to a community, do these models exhibit community-aligned behavior that generalizes to novel events and ambiguous information, or do they merely reproduce memorized facts~\cite{li2023steerability, bisbee2024synthetic}? In other words, do they act in ways consistent with how the community itself would act when confronted with new uncertainty, rather than simply replaying past reactions? The answer carries significant implications for both research and safety. If outputs are driven by shallow mimicry, simulations and forecasts built on these models cannot be trusted when applied to new or unforeseen situations. Reliable simulation and decision-support applications require models that capture the deeper behavioral patterns---structured, probabilistic tendencies for responding under epistemic uncertainty---that govern how a community reacts.

At the same time, if alignment encodes these behavioral patterns, it also embeds the biases and blind spots of those communities. Without proper evaluation, these biases may be amplified or misrepresented in downstream applications, including those in policy, risk analysis, or safety-critical systems. Understanding whether models generalize community behavior or simply replay surface patterns is therefore essential for both leveraging aligned LLMs in simulation and mitigating their risks.

Current validation methods often blur the line between true stance transfer and simple memorization. Because they rarely test models on genuinely novel scenarios, evaluations can mistake surface-level mimicry for stable, community-specific behavioral patterns, or conflate factual recall with transferable strategies for handling uncertainty~\cite{prpa2024challenges, li2024personal, jiang2024investigating}. This limitation is especially critical in high-stakes settings where reliable behavior under epistemic uncertainty is essential.

To address this limitation, we introduce a pipeline that aligns LLMs to specific communities and applies targeted knowledge deletion to remove factual recall of events, thereby enforcing a state of epistemic uncertainty. We then evaluate whether models maintain community-specific behavioral patterns when confronted with novel scenarios, using Russian and Ukrainian military blogger discourse and American political communities as testbeds. This setup ensures that any consistent patterns observed cannot be explained by factual recall alone, providing controlled evidence of epistemic stance transfer: even after aggressive fact removal, aligned models demonstrate stable, community-consistent behaviors under uncertainty.

This persistence shows that alignment encodes durable, community-specific behavioral patterns that operate independently of factual knowledge. For simulation and decision-support applications, this is promising: aligned LLMs appear capable of generalizing these behaviors beyond known events. At the same time, these findings highlight the need for evaluation frameworks that probe structured behavioral patterns directly, rather than relying solely on surface-level accuracy.

\begin{figure*}[t]

\includegraphics[width=1\textwidth]{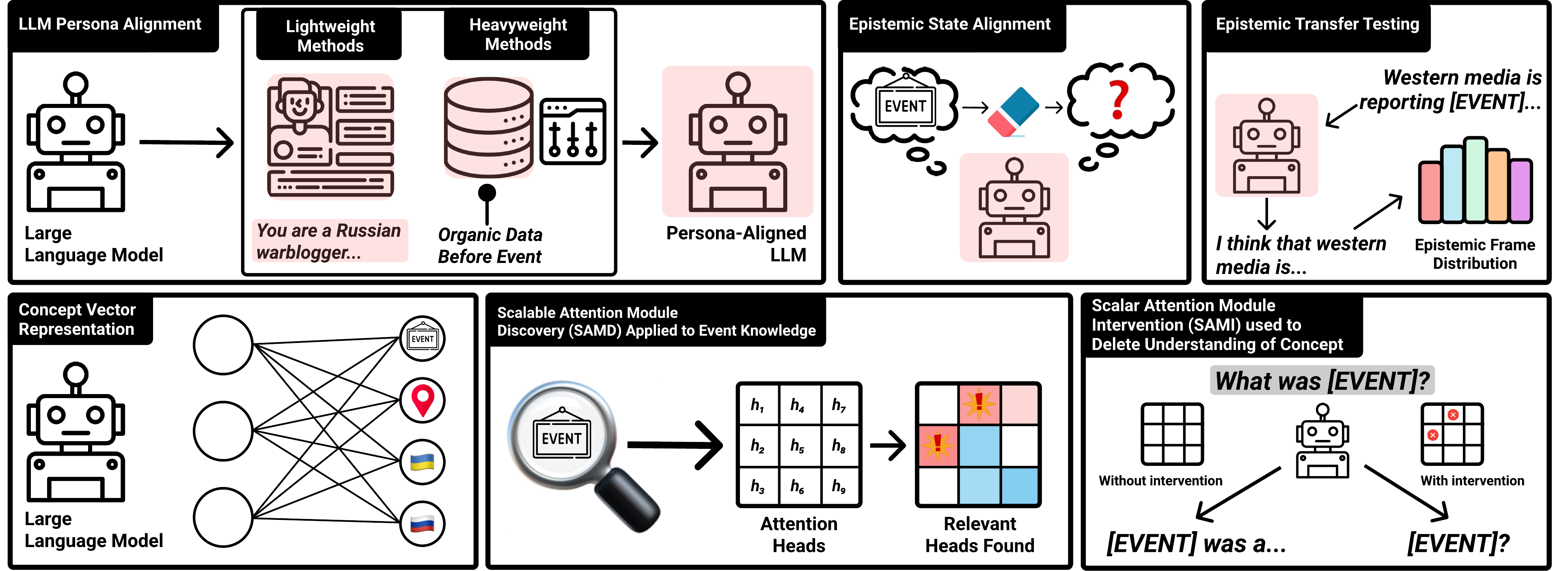}
\caption{\textbf{Epistemic Stance Transfer Testing.} This figure illustrates how we evaluate whether LLMs exhibit stable, community-aligned \emph{behavioral patterns under uncertainty} or simply rely on factual recall. \textit{Top row:} Models are aligned to a community using lightweight methods (e.g., system prompts, prepended examples) or heavyweight methods (fine-tuning on pre-event discourse), then have knowledge of key events deleted via SAMI to simulate pre-event ignorance. These knowledge-deleted models are then evaluated on novel scenarios. \textit{Bottom row:} The knowledge deletion method is adapted from Su et al.~\cite{su2025concepts}, and the visualization style is inspired by their work.}
    \label{fig:pipeline}
\end{figure*}



\section{Methodology}

We ground our approach in epistemic stance transfer, where communities develop systematic behavioral patterns for responding under epistemic uncertainty: contexts where knowledge is incomplete or contested~\cite{li2023steerability, goldman1999knowledge, fricker2007epistemic}. Rather than simply assigning different probabilities to the same events, communities adopt \emph{epistemic stances} that shape how evidence is represented, updated, deferred, or acted upon under ambiguity. Formally, we model a community stance as a stochastic mapping over response operations,\[
\Psi_c:\ \mathcal{E}\times\mathcal{U}\ \to\ \Delta(\mathcal{R}),\qquad r\sim \Psi_c(e,u),
\]

where \(\mathcal{R}=\{r_1,\dots,r_K\}\) is a fixed set of response operations (e.g., \textit{suspend\_judgement}, \textit{dispute}, \textit{seek\_alternative\_cause}, \textit{demand\_evidence}, \textit{defer\_to\_authority}), \(\mathcal{E}\) is event content, and \(\mathcal{U}\) is an uncertainty context. The organic community induces a distribution \(\pi_c(\cdot\mid u)\in\Delta(\mathcal{R})\) by conditioning on \(u\) and marginalizing over event content. An aligned model with parameters \(\theta\) induces \(q^{\text{env}}_\theta(\cdot\mid u,c)\) under two environments, facts present and targeted deletion of event knowledge. Our evaluation tests whether the model’s distribution over response operations \emph{matches the organic distribution} and whether this alignment remains \emph{invariant} to knowledge deletion:

\[
D\!\left(q^{\text{env}}_\theta(\cdot\mid u,c),\ \pi_c(\cdot\mid u)\right)\ \text{is small for env}\in\{\text{facts},\text{deleted}\},
\]

with \(D\) given by Jensen–Shannon divergence~\cite{menendez1997jensen}. For example, let \(\mathcal{R}=\{\texttt{dispute},\ \texttt{alt\_cause},\ \texttt{suspend\_judgement},\ \texttt{demand\_evidence},\ \texttt{defer}\}\). The organic Russian military-blogger distribution \(\pi_{RU}(\cdot\mid u)\) places relatively higher mass on \textit{dispute} and \textit{alt\_cause}, while the Ukrainian distribution \(\pi_{UA}(\cdot\mid u)\) places relatively higher mass on \textit{suspend\_judgement} and \textit{demand\_evidence}. The question is not whether Russian discourse equals a single label, but whether
\[
q^{\text{deleted}}_\theta(\cdot\mid u,\text{RU})\ \text{is close to}\ \pi_{\text{RU}}(\cdot\mid u)\quad
\]
and whether this holds across contexts when factual recall is suppressed. We list full definitions, identification claims, hypotheses, and estimation details in Appendix~\ref{app:theory}.


\subsection{Community Discourse Corpora}
We test epistemic stance transfer using discourse from oppositional community pairs: Russian versus Ukrainian military bloggers on Telegram during the ongoing conflict, and conservative versus liberal users on Twitter/X during the 2024 U.S. presidential election~\cite{balasubramanian2024public, gerard2025bridging}. 

These corpora offer two key advantages for testing epistemic stance transfer. First, they feature communities with fundamentally different epistemic approaches to identical events~\cite{gerard2025fear, gerard2025modeling, oleinik2024telegram}. Second, their temporal structure allows for controlled knowledge deletion: we can align models on pre-event discourse, delete all knowledge of target events, then test responses to event-related scenarios.



\textbf{Frame Identification and Annotation}: For each target event, domain experts use systematic discourse analysis to identify naturally occurring epistemic response strategies in community responses~\cite{terry2017thematic}, emphasizing deeper patterns in evidence evaluation, source credibility, and causal attribution rather than surface-level linguistic cues. For example, for the Bucha event (see Appendix~\ref{app:corpus} for more details on Bucha), Russian discourse is driven by conspiratorial skepticism while Ukrainian discourse is dominated by evidentiary realism and moral absolutism frameworks. These frames capture fundamental differences in epistemic approaches to novel information. Annotation procedures follow established protocols with two domain expert coders achieving high inter-annotator agreement ($\kappa$ = 0.76-0.84 across events). Full frame taxonomies and guidelines appear in Appendix~\ref{app:corpus}.

\subsection{Community Alignment}
We test six alignment conditions that span the methodological spectrum of persona-based alignment techniques, which we categorize as lightweight and heavyweight based on computational complexity (implementation details in Appendix~\ref{app:finetuning}). These alignment approaches adapt established persona-based techniques that range from simple prompt engineering to intensive fine-tuning~\cite{schuller2024generating}. 


\textbf{Lightweight methods} rely on prompt engineering without parameter updates, making them accessible and computationally efficient. \textit{System-Prompt} alignment uses community role instructions that establish the model's persona and perspective (e.g., ``You are a Russian war blogger analyzing events from a pro-Russian perspective''). \textit{Prepended} alignment primes models by prepending community-role instructions directly to each input prompt. These approaches follow standard practices in persona-based modeling but require no training data beyond inference~\cite{schuller2024generating}.

\textbf{Heavyweight methods} require supervised fine-tuning and are typically employed when lightweight approaches cannot sufficiently capture complex community behaviors or when working with niche communities~\cite{he2024community, 10.1145/3613905.3651116}. Our \textit{Finetuned} conditions use supervised fine-tuning on authentic community discourse collected prior to target events; we follow established community alignment practices~\cite{he2024community} for this. This approach embeds community-specific patterns directly into model parameters rather than relying on prompt-level guidance.

\textbf{Baseline conditions} are used to establish experimental controls to validate epistemic stance specificity. \textit{Oracle} conditions retain full event knowledge to establish performance baselines (i.e., no targeted knowledge deletion takes place), while \textit{Cross-Community} conditions test specificity by evaluating models aligned with one community against opposing community patterns (e.g., Ukrainian-aligned models evaluated against Russian discourse frameworks). \textit{Vanilla} conditions apply knowledge deletion to unaligned models. 


All conditions except Oracle undergo targeted knowledge deletion prior to evaluation to ensure that observed community-specific patterns reflect learned behavioral patterns under uncertainty rather than mere factual recall of target events.

\subsection{Inducing Epistemic Uncertainty via Unlearning}
After aligning an LLM to a specific community, we enforce complete ignorance of target events to isolate \emph{epistemic stance transfer} from factual recall. Because these events were likely encountered during pretraining~\cite{jiang2024investigating}, this intervention distinguishes genuine community-aligned behavioral patterns from simple retrieval. Consistent stance distributions after deletion indicate reliance on generalized response mappings rather than event-specific recall.

We apply Scalable Attention Module Discovery and Scalar Attention Module Intervention (SAMD/SAMI)~\cite{su2025concepts} to suppress event-specific knowledge while preserving general capabilities. SAMD ranks attention modules by cosine similarity between event concept vectors and module outputs, and SAMI attenuates those modules’ contributions during inference, creating a controlled dissociation between factual recall and community-aligned behavioral patterns (see Appendix~\ref{app:sami_details}). We validate SAMI with three established evaluation approaches: structured factual recall (LAMA-style cloze tasks~\cite{petroni2019language}), direct information probing, and paraphrasing/adversarial recovery~\cite{jiang2020can} (see Appendix~\ref{app:sami_validation}). Post-unlearning models resist knowledge recovery while maintaining unrelated task performance, ensuring that observed behavioral patterns reflect generalized alignment rather than memorized facts.



\subsection{Measuring Framework Transfer via Discourse Classification}
After knowledge deletion, we prompt models with snippets from organic community discussions of target events. To control for leakage and bias, we implement two safeguards. First, we anonymize event details to reduce the chance of residual memorization (e.g., for the Bucha event, prompting with ``Western media is reporting alleged crimes committed by Russian troops in a Ukrainian town'' instead of naming the location). Although results are similar with non-anonymized prompts, anonymization adds an extra layer of protection. Second, we remove prompts that contain leading indicators that could bias responses toward particular interpretations. This setup helps ensure that any community-specific patterns in model outputs stem from internalized epistemic stances rather than factual recall. 
Dataset statistics are shown in Appendix~\ref{app:corpus}.

\subsection{Evaluating Epistemic Stance Transfer}
For each event setup, we train classifiers on human-validated stance annotations and rigorously validate their performance on both organic community discourse and model-generated responses. This dual validation helps ensure that the classifiers capture epistemic stance patterns consistently across both human and model-generated discourse. Full details of the training and validation procedures are provided in Appendix~\ref{app:classifier}. Using these classifiers, we evaluate stance transfer at two levels: (1) macro-level distributional similarity, and (2) individual-level coherence under uncertainty.

\textbf{Measuring Epistemic Stance Transfer}: We quantify framework transfer by comparing model response distributions to organic community baselines using the Jensen–Shannon (JS) divergence~\cite{menendez1997jensen}, which selected for its symmetry, interpretability, and bounded [0,1] range. While we also evaluated four alternative divergence measures, their outputs were highly correlated with JS (Appendix~\ref{app:full-exp}), indicating that our findings hold regardless of the metric used. Organic community discourse serves as the ground truth for authentic epistemic stances, reflecting how communities process ambiguous information in real-world contexts. Models that transfer these stances should generate response distributions that closely align with these patterns, where smaller distributional divergence signal stronger framework transfer.

\textbf{Measuring Epistemic Stance Coherence}: Beyond distributional similarity, we assess consistency of epistemic stance application under uncertainty. For each test scenario, we generate $N=5$ independent completions and compute epistemic stance entropy: $H(p_i) = -\sum_{k=1}^K p_{i,k} \log_2 p_{i,k}$, where $p_{i,k}$ represents the frequency of stance $k$ across generations. Low entropy ($H \approx 0$) indicates consistent application of the same epistemic stance under genuine ignorance, while high entropy ($H \approx \log_2 K$) indicates policy instability and inconsistent response patterns. This distinguishes genuine epistemic stance transfer from superficial pattern matching: models with stable, community-aligned behavioral patterns maintain low entropy even after knowledge deletion, while models reliant on factual recall exhibit sharp entropy increases when deprived of those facts.

\pgfplotsset{
    colormap={cool}{rgb255(0cm)=(255,255,255); rgb255(1cm)=(0,128,255); rgb255(2cm)=(255,0,255)}
}
\pgfplotsset{
    colormap={violet1}{color=(white) rgb255=(238,140,238) rgb255=(25,25,122)}
}

\pgfplotsset{
    colormap={violet1}{color=(white) rgb255=(238,140,238) rgb255=(25,25,122)}
}

\pgfplotsset{
    colormap={violet1}{color=(white) rgb255=(238,140,238) rgb255=(25,25,122)}
}

\pgfplotsset{
  colormap={violet1}{
    rgb=(1,1,1)
    rgb255=(238,140,238)
    rgb255=(25,25,122)
  }
}

\begin{figure}[t]
\centering
\begin{minipage}[t]{0.49\textwidth}
\centering
\resizebox{\textwidth}{!}{
\begin{tikzpicture}
\begin{axis}[
    view={0}{90},
    colorbar,
    colorbar style={
        ylabel={Jensen-Shannon Divergence},
        ylabel style={rotate=180}
    },
    xlabel={Model Condition},
    ylabel={Model Condition},
    xticklabels={
        Organic,
        Oracle,
        Finetuned,
        System Prompt,
        Prepend,
        Cross-Community,
        Vanilla
    },
    yticklabels={
        Organic,
        Oracle,
        Finetuned,
        System Prompt,
        Prepend,
        Cross-Community,
        Vanilla
    },
    xtick={0,1,2,3,4,5,6},
    ytick={0,1,2,3,4,5,6},
    x tick label style={rotate=90, anchor=north east},
    y tick label style={anchor=east},
    enlargelimits=false,
    axis equal image,
    width=12cm,
    height=10cm,
    point meta min=0,
    point meta max=0.5,
]
\addplot3[
    matrix plot,
    mesh/cols=7,
    point meta=explicit
] coordinates {
    (0,0,0) [0.0]
    (1,0,0) [0.08]
    (2,0,0) [0.11363205409475488]
    (3,0,0) [0.14434816306140652]
    (4,0,0) [0.16]
    (5,0,0) [0.2864983816992409]
    (6,0,0) [0.4574043432304754]
    
    (0,1,0) [0.08]
    (1,1,0) [0.0]
    (2,1,0) [0.10]
    (3,1,0) [0.16]
    (4,1,0) [0.14]
    (5,1,0) [0.28]
    (6,1,0) [0.45]
    
    (0,2,0) [0.11363205409475488]
    (1,2,0) [0.10]
    (2,2,0) [0.0]
    (3,2,0) [0.21783779139781684]
    (4,2,0) [0.19]
    (5,2,0) [0.24690643366667628]
    (6,2,0) [0.44425700829272974]
    
    (0,3,0) [0.14434816306140652]
    (1,3,0) [0.16]
    (2,3,0) [0.21783779139781684]
    (3,3,0) [0.0]
    (4,3,0) [0.10]
    (5,3,0) [0.32283561734306204]
    (6,3,0) [0.4378748118026829]
    
    (0,4,0) [0.16]
    (1,4,0) [0.14]
    (2,4,0) [0.19]
    (3,4,0) [0.10]
    (4,4,0) [0.0]
    (5,4,0) [0.26]
    (6,4,0) [0.42]
    
    (0,5,0) [0.2864983816992409]
    (1,5,0) [0.28]
    (2,5,0) [0.24690643366667628]
    (3,5,0) [0.32283561734306204]
    (4,5,0) [0.26]
    (5,5,0) [0.0]
    (6,5,0) [0.23740153609823444]
    
    (0,6,0) [0.4574043432304754]
    (1,6,0) [0.45]
    (2,6,0) [0.44425700829272974]
    (3,6,0) [0.4378748118026829]
    (4,6,0) [0.42]
    (5,6,0) [0.23740153609823444]
    (6,6,0) [0.0]
};

\node at (axis cs:0,0,0) { 0.00};
\node at (axis cs:1,0,0) { \begin{tabular}{c} 0.08 \\ *** \end{tabular}};
\node at (axis cs:2,0,0) { \begin{tabular}{c} 0.11 \\ *** \end{tabular}};
\node at (axis cs:3,0,0) { \begin{tabular}{c} 0.15 \\ *** \end{tabular}};
\node at (axis cs:4,0,0) { \begin{tabular}{c} 0.16 \\ *** \end{tabular}};
\node at (axis cs:5,0,0) { \begin{tabular}{c} 0.29 \\ *** \end{tabular}};
\node at (axis cs:6,0,0) { \begin{tabular}{c} 0.46 \\ *** \end{tabular}};

\node at (axis cs:0,1,0) { \begin{tabular}{c} 0.08 \\ *** \end{tabular}};
\node at (axis cs:1,1,0) { 0.00};
\node at (axis cs:2,1,0) { \begin{tabular}{c} 0.10 \\ *** \end{tabular}};
\node at (axis cs:3,1,0) { \begin{tabular}{c} 0.16 \\ *** \end{tabular}};
\node at (axis cs:4,1,0) { \begin{tabular}{c} 0.14 \\ *** \end{tabular}};
\node at (axis cs:5,1,0) { \begin{tabular}{c} 0.28 \\ *** \end{tabular}};
\node at (axis cs:6,1,0) { \begin{tabular}{c} 0.45 \\ *** \end{tabular}};

\node at (axis cs:0,2,0) { \begin{tabular}{c} 0.11 \\ *** \end{tabular}};
\node at (axis cs:1,2,0) { \begin{tabular}{c} 0.10 \\ *** \end{tabular}};
\node at (axis cs:2,2,0) { 0.00};
\node at (axis cs:3,2,0) { \begin{tabular}{c} 0.22 \\ *** \end{tabular}};
\node at (axis cs:4,2,0) { \begin{tabular}{c} 0.19 \\ *** \end{tabular}};
\node at (axis cs:5,2,0) { \begin{tabular}{c} 0.25 \\ *** \end{tabular}};
\node at (axis cs:6,2,0) { \begin{tabular}{c} 0.44 \\ *** \end{tabular}};

\node at (axis cs:0,3,0) { \begin{tabular}{c} 0.15 \\ *** \end{tabular}};
\node at (axis cs:1,3,0) { \begin{tabular}{c} 0.16 \\ *** \end{tabular}};
\node at (axis cs:2,3,0) { \begin{tabular}{c} 0.22 \\ *** \end{tabular}};
\node at (axis cs:3,3,0) { 0.00};
\node at (axis cs:4,3,0) { \begin{tabular}{c} 0.10 \\ *** \end{tabular}};
\node at (axis cs:5,3,0) { \begin{tabular}{c} 0.32 \\ *** \end{tabular}};
\node at (axis cs:6,3,0) { \begin{tabular}{c} 0.44 \\ *** \end{tabular}};

\node at (axis cs:0,4,0) { \begin{tabular}{c} 0.16 \\ *** \end{tabular}};
\node at (axis cs:1,4,0) { \begin{tabular}{c} 0.14 \\ *** \end{tabular}};
\node at (axis cs:2,4,0) { \begin{tabular}{c} 0.19 \\ *** \end{tabular}};
\node at (axis cs:3,4,0) { \begin{tabular}{c} 0.10 \\ *** \end{tabular}};
\node at (axis cs:4,4,0) { 0.00};
\node at (axis cs:5,4,0) { \begin{tabular}{c} 0.26 \\ *** \end{tabular}};
\node at (axis cs:6,4,0) { \begin{tabular}{c} 0.42 \\ *** \end{tabular}};

\node at (axis cs:0,5,0) { \begin{tabular}{c} 0.29 \\ *** \end{tabular}};
\node at (axis cs:1,5,0) { \begin{tabular}{c} 0.28 \\ *** \end{tabular}};
\node at (axis cs:2,5,0) { \begin{tabular}{c} 0.25 \\ *** \end{tabular}};
\node at (axis cs:3,5,0) { \begin{tabular}{c} 0.32 \\ *** \end{tabular}};
\node at (axis cs:4,5,0) { \begin{tabular}{c} 0.26 \\ *** \end{tabular}};
\node at (axis cs:5,5,0) { 0.00};
\node at (axis cs:6,5,0) { \begin{tabular}{c} 0.24 \\ *** \end{tabular}};

\node at (axis cs:0,6,0) { \begin{tabular}{c} 0.46 \\ *** \end{tabular}};
\node at (axis cs:1,6,0) { \begin{tabular}{c} 0.45 \\ *** \end{tabular}};
\node at (axis cs:2,6,0) { \begin{tabular}{c} 0.44 \\ *** \end{tabular}};
\node at (axis cs:3,6,0) { \begin{tabular}{c} 0.44 \\ *** \end{tabular}};
\node at (axis cs:4,6,0) { \begin{tabular}{c} 0.42 \\ *** \end{tabular}};
\node at (axis cs:5,6,0) { \begin{tabular}{c} 0.24 \\ *** \end{tabular}};
\node at (axis cs:6,6,0) { 0.00};

\end{axis}
\end{tikzpicture}
}
\subcaption{Russian Community Baseline}
\label{fig:russian_baseline}
\end{minipage}
\hfill
\begin{minipage}[t]{0.47\textwidth}
\centering
\resizebox{\textwidth}{!}{
\begin{tikzpicture}
\begin{axis}[
    view={0}{90},
    colorbar,
    colorbar style={
        ylabel={Jensen-Shannon Divergence},
        ylabel style={rotate=180}
    },
    xlabel={Model Condition},
    xticklabels={
        Organic,
        Oracle,
        Finetuned,
        System Prompt,
        Prepend,
        Cross-Community,
        Vanilla
    },
    yticklabels={
        Organic,
        Oracle,
        Finetuned,
        System Prompt,
        Prepend,
        Cross-Community,
        Vanilla
    },
    xtick={0,1,2,3,4,5,6},
    ytick={0,1,2,3,4,5,6},
    x tick label style={rotate=90, anchor=north east},
    y tick label style={anchor=east},
    enlargelimits=false,
    axis equal image,
    width=12cm,
    height=10cm,
    point meta min=0,
    point meta max=0.5,
]
\addplot3[
    matrix plot,
    mesh/cols=7,
    point meta=explicit
] coordinates {
    (0,0,0) [0.0]
    (1,0,0) [0.07]
    (2,0,0) [0.11]
    (3,0,0) [0.12]
    (4,0,0) [0.11]
    (5,0,0) [0.31]
    (6,0,0) [0.48]
    
    (0,1,0) [0.07]
    (1,1,0) [0.0]
    (2,1,0) [0.08]
    (3,1,0) [0.11]
    (4,1,0) [0.13]
    (5,1,0) [0.30]
    (6,1,0) [0.45]
    
    (0,2,0) [0.11]
    (1,2,0) [0.08]
    (2,2,0) [0.0]
    (3,2,0) [0.20]
    (4,2,0) [0.21]
    (5,2,0) [0.27]
    (6,2,0) [0.45]
    
    (0,3,0) [0.12]
    (1,3,0) [0.11]
    (2,3,0) [0.20]
    (3,3,0) [0.0]
    (4,3,0) [0.11]
    (5,3,0) [0.35]
    (6,3,0) [0.46]
    
    (0,4,0) [0.11]
    (1,4,0) [0.13]
    (2,4,0) [0.21]
    (3,4,0) [0.11]
    (4,4,0) [0.0]
    (5,4,0) [0.29]
    (6,4,0) [0.44]
    
    (0,5,0) [0.31]
    (1,5,0) [0.30]
    (2,5,0) [0.27]
    (3,5,0) [0.35]
    (4,5,0) [0.29]
    (5,5,0) [0.0]
    (6,5,0) [0.22]
    
    (0,6,0) [0.48]
    (1,6,0) [0.45]
    (2,6,0) [0.45]
    (3,6,0) [0.46]
    (4,6,0) [0.44]
    (5,6,0) [0.22]
    (6,6,0) [0.0]
};

\node at (axis cs:0,0,0) { 0.00};
\node at (axis cs:1,0,0) { \begin{tabular}{c} 0.07 \\ *** \end{tabular}};
\node at (axis cs:2,0,0) { \begin{tabular}{c} 0.11 \\ *** \end{tabular}};
\node at (axis cs:3,0,0) { \begin{tabular}{c} 0.12 \\ *** \end{tabular}};
\node at (axis cs:4,0,0) { \begin{tabular}{c} 0.11 \\ *** \end{tabular}};
\node at (axis cs:5,0,0) { \begin{tabular}{c} 0.31 \\ *** \end{tabular}};
\node at (axis cs:6,0,0) { \begin{tabular}{c} 0.48 \\ *** \end{tabular}};

\node at (axis cs:0,1,0) { \begin{tabular}{c} 0.07 \\ *** \end{tabular}};
\node at (axis cs:1,1,0) { 0.00};
\node at (axis cs:2,1,0) { \begin{tabular}{c} 0.08 \\ *** \end{tabular}};
\node at (axis cs:3,1,0) { \begin{tabular}{c} 0.11 \\ *** \end{tabular}};
\node at (axis cs:4,1,0) { \begin{tabular}{c} 0.13 \\ *** \end{tabular}};
\node at (axis cs:5,1,0) { \begin{tabular}{c} 0.30 \\ *** \end{tabular}};
\node at (axis cs:6,1,0) { \begin{tabular}{c} 0.45 \\ *** \end{tabular}};

\node at (axis cs:0,2,0) { \begin{tabular}{c} 0.11 \\ *** \end{tabular}};
\node at (axis cs:1,2,0) { \begin{tabular}{c} 0.08 \\ *** \end{tabular}};
\node at (axis cs:2,2,0) { 0.00};
\node at (axis cs:3,2,0) { \begin{tabular}{c} 0.20 \\ *** \end{tabular}};
\node at (axis cs:4,2,0) { \begin{tabular}{c} 0.21 \\ *** \end{tabular}};
\node at (axis cs:5,2,0) { \begin{tabular}{c} 0.27 \\ *** \end{tabular}};
\node at (axis cs:6,2,0) { \begin{tabular}{c} 0.45 \\ *** \end{tabular}};

\node at (axis cs:0,3,0) { \begin{tabular}{c} 0.12 \\ *** \end{tabular}};
\node at (axis cs:1,3,0) { \begin{tabular}{c} 0.11 \\ *** \end{tabular}};
\node at (axis cs:2,3,0) { \begin{tabular}{c} 0.20 \\ *** \end{tabular}};
\node at (axis cs:3,3,0) { 0.00};
\node at (axis cs:4,3,0) { \begin{tabular}{c} 0.11 \\ *** \end{tabular}};
\node at (axis cs:5,3,0) { \begin{tabular}{c} 0.35 \\ *** \end{tabular}};
\node at (axis cs:6,3,0) { \begin{tabular}{c} 0.46 \\ *** \end{tabular}};

\node at (axis cs:0,4,0) { \begin{tabular}{c} 0.11 \\ *** \end{tabular}};
\node at (axis cs:1,4,0) { \begin{tabular}{c} 0.13 \\ *** \end{tabular}};
\node at (axis cs:2,4,0) { \begin{tabular}{c} 0.21 \\ *** \end{tabular}};
\node at (axis cs:3,4,0) { \begin{tabular}{c} 0.11 \\ *** \end{tabular}};
\node at (axis cs:4,4,0) { 0.00};
\node at (axis cs:5,4,0) { \begin{tabular}{c} 0.29 \\ *** \end{tabular}};
\node at (axis cs:6,4,0) { \begin{tabular}{c} 0.44 \\ *** \end{tabular}};

\node at (axis cs:0,5,0) { \begin{tabular}{c} 0.31 \\ *** \end{tabular}};
\node at (axis cs:1,5,0) { \begin{tabular}{c} 0.30 \\ *** \end{tabular}};
\node at (axis cs:2,5,0) { \begin{tabular}{c} 0.27 \\ *** \end{tabular}};
\node at (axis cs:3,5,0) { \begin{tabular}{c} 0.35 \\ *** \end{tabular}};
\node at (axis cs:4,5,0) { \begin{tabular}{c} 0.29 \\ *** \end{tabular}};
\node at (axis cs:5,5,0) { 0.00};
\node at (axis cs:6,5,0) { \begin{tabular}{c} 0.22 \\ *** \end{tabular}};

\node at (axis cs:0,6,0) { \begin{tabular}{c} 0.48 \\ *** \end{tabular}};
\node at (axis cs:1,6,0) { \begin{tabular}{c} 0.45 \\ *** \end{tabular}};
\node at (axis cs:2,6,0) { \begin{tabular}{c} 0.45 \\ *** \end{tabular}};
\node at (axis cs:3,6,0) { \begin{tabular}{c} 0.46 \\ *** \end{tabular}};
\node at (axis cs:4,6,0) { \begin{tabular}{c} 0.44 \\ *** \end{tabular}};
\node at (axis cs:5,6,0) { \begin{tabular}{c} 0.22 \\ *** \end{tabular}};
\node at (axis cs:6,6,0) { 0.00};

\end{axis}
\end{tikzpicture}
}
\subcaption{Ukrainian Community Baseline}
\label{fig:ukrainian_baseline}
\end{minipage}

\caption{Jensen-Shannon divergence matrices comparing model conditions against organic community baselines for Bucha massacre discourse. White cells indicate perfect similarity (divergence = 0), darker cells greater divergence. Both Russian-aligned (a) and Ukrainian-aligned (b) models cluster closely to organic baselines (Oracle, Finetuned, System Prompt, Prepend), while Cross-Community and Vanilla conditions deviate. Statistical significance (***: p < 0.01) was determined via permutation testing with 1000 iterations. The pattern demonstrates robust epistemic stance transfer across alignment methods regardless of factual knowledge deletion.}
\label{fig:distance_matrices}
\end{figure}
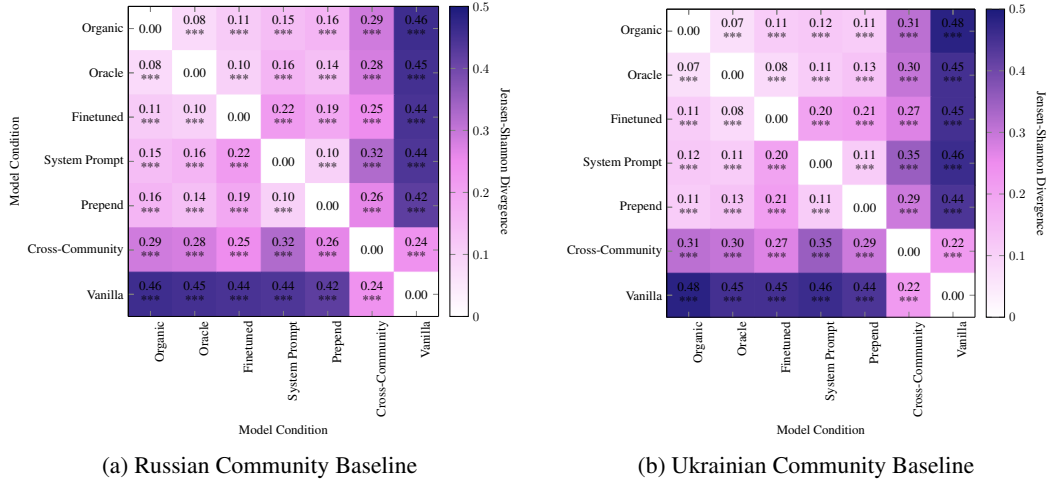
\vspace{-2mm}

\begin{figure}[ht]
\centering
\begin{tikzpicture}[scale=0.6]
  \begin{axis}[
    title={Macro Jensen-Shannon Divergence from Organic Baselines},
    xbar,
    width=10cm,
    xlabel={Jensen-Shannon Divergence},
    bar width=18pt,
    y axis line style={opacity=1},
    x axis line style={opacity=1},
axis x line = bottom,
axis y line = left,
    tickwidth=0pt,
    ytick=data,
    xmax=0.5,
    enlarge y limits=0.2,
    enlarge x limits=0.02,
    symbolic y coords={Oracle,Finetuned,System Prompt, Prepend, Cross-Community, Vanilla},
    nodes near coords,
    nodes near coords align={horizontal},
    every node near coord/.append style={
      xshift=10pt,
      black,
      /pgf/number format/fixed,
      /pgf/number format/precision=2,
    },
    minor x tick num=1,
    tick label style={black},
    title style={black},
    label style={black},
    xticklabel style={/pgf/number format/fixed,/pgf/number format/precision=2},
  ]
    \addplot+[
      draw=black,
      fill=blue!30,
      pattern=crosshatch,
      pattern color=blue!30,
      error bars/.cd,
        x dir=both,
        x explicit,
    ] coordinates {
      (0.077,Oracle)          +- (0.005,0)
      (0.107,Finetuned)       +- (0.006,0)
      (0.127,System Prompt)   +- (0.012,0)
      (0.132,Prepend)         +- (0.008,0)
      (0.300,Cross-Community) +- (0.017,0)
      (0.435,Vanilla)         +- (0.020,0)
    };
  \end{axis}
\end{tikzpicture}
\caption{Macro average Jensen-Shannon divergence from organic baselines across all events and communities. Aligned conditions cluster at low divergences, while misaligned conditions show substantially higher divergences (0.300–0.435), confirming community-specific epistemic stance transfer. Error bars represent standard error across six event–community combinations.}
\label{fig:macro_js_distance}
\end{figure}
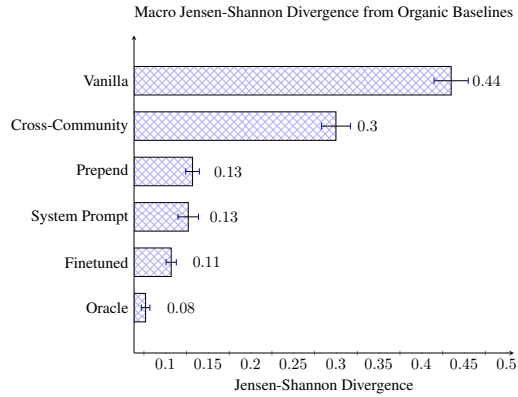

\section{Results}
Our experimental design tests whether LLMs have internalized how a community responds under uncertainty to new situations, or if they simply echo past reactions without deeper behavioral alignment. We compare three hypotheses: \textit{memorization} (knowledge deletion should sharply degrade performance), \textit{surface mimicry} (lightweight prompting without facts should fail to reproduce consistent behavioral patterns), and \textit{behavioral transfer} (aligned models should preserve stable uncertainty-response patterns even after knowledge deletion).


Strong evidence supports framework transfer. As shown in Figures~\ref{fig:distance_matrices} and~\ref{fig:macro_js_distance}, community-aligned models cluster closely to organic baselines even after complete factual knowledge removal, whereas misaligned/unaligned models show systematic deviations. At the \textbf{macro level}, aligned models produce stance distributions that closely match organic baselines, indicating that high-level behavioral patterns persist even after event-specific knowledge is removed. At the \textbf{micro level}, aligned models maintain significantly lower entropy ($H < 0.7$ bits) than controls ($H > 1.2$ bits, $p < 0.001$), demonstrating stable, low-variance behavioral patterns under uncertainty rather than random or brittle outputs. Together, these results suggest that epistemic stance transfer is structural rather than factual, reflecting generalizable, community-aligned behavioral patterns rather than simple factual recall.

\section{Limitations and Ethical Considerations}
\textbf{Limitations}: SAMI successfully removes explicit factual recall about target events, but complete knowledge elimination across all model components cannot be guaranteed. Additionally, our study focuses on temporally bounded events (military conflicts, political debates) which allow for clean pre/post-event data separation, but this may limit generalizability to other epistemic domains. Testing on non-temporal epistemic communities (e.g., scientific forums, academic disciplines) would strengthen claims about epistemic stance transfer across diverse knowledge domains, and we plan to extend our methodology to such contexts in future work.

\textbf{Ethical Considerations}: This research carries dual-use risks that require careful scrutiny. While our methods are designed to support AI safety by detecting and mitigating epistemic biases, the same techniques could be misused to amplify harmful narratives by deliberately inducing persistent behavioral patterns through targeted prompting or fine-tuning. The relative ease with which models adopt community-specific behavioral patterns raises concerns about inadvertent bias propagation in deployed systems, particularly when models are trained on data from specific communities or ideological groups. To mitigate these risks, we focus our contributions on detection and evaluation methodologies rather than optimization techniques for bias induction, restrict access to sensitive community discourse data, and emphasize the development of safeguards to identify and counter epistemic manipulation in real-world deployments.

\section{Implications and Future Work}

Our results suggest that models can encode deep, systematic uncertainty-processing biases even when all factual knowledge is stripped away. In the context of AI safety, this means that filtering based solely on surface-level outputs or factual correctness risks missing persistent, harder-to-detect biases that can reemerge in novel scenarios. Safety evaluations must therefore extend beyond accuracy metrics to probe the community-aligned behavioral patterns that models carry into new domains. For epistemic AI, however, these findings are promising: LLMs exhibit stable, community-aligned behavioral patterns, applying them consistently under conditions of ignorance.


We are extending our evaluation to medical, legal, and scientific domains to assess the generalizability of epistemic stance transfer beyond political discourse. Additionally, we plan to incorporate mechanistic interpretability techniques (e.g., Su et al.~\cite{su2025concepts}) to pinpoint when robust stance transfer emerges during fine-tuning. Longer term, our goal is to develop tools for detecting epistemic biases in deployed systems and to investigate the internal model pathways that drive framework transfer.


\section{Conclusions}

Aligned LLMs may acquire community-aligned epistemic stances that persist even when factual knowledge is removed. Through targeted knowledge deletion, we show that these models reproduce systematic stance patterns under ignorance and that such patterns can be elicited through simple prompting. These findings challenge safety frameworks that focus solely on output accuracy while overlooking deeper, behavioral biases in how models process uncertainty. Our work provides a foundation for auditing aligned systems not only for factual accuracy but also for the epistemic biases that shape their behavior, a necessary step toward safer and more accountable LLM deployments.

\clearpage
\bibliography{neurips}
\clearpage

\appendix

\section{Theoretical grounding and identification}
\label{app:theory}

\subsection{Stances as stochastic mappings over response operations}
Let \(\mathcal{R}=\{r_1,\dots,r_K\}\) denote response operations such as \textit{suspend\_judgement}, \textit{dispute}, \textit{seek\_alternative\_cause}, \textit{demand\_evidence}, and \textit{defer\_to\_authority}. Let \(\mathcal{E}\) be event content and \(\mathcal{U}\) be uncertainty contexts. For community \(c\) an epistemic stance is formalized as a stochastic mapping over these operations:
\[
\Psi_c:\ \mathcal{E}\times\mathcal{U}\ \to\ \Delta(\mathcal{R}),\qquad r\sim \Psi_c(e,u),
\]
where \(\Delta(\mathcal{R})\) is the probability simplex on \(\mathcal{R}\). A stance yields a distribution on \(\mathcal{R}\), not a single outcome.

We also view a stance as a structured object
\[
\Psi_c=\langle K_c,\ U_c,\ D_c,\ T_c\rangle,
\]
where \(K_c\) is a belief state representation, \(U_c\) is an evidence update rule, \(D_c\) is a decision criterion under ambiguity, and \(T_c\) is a stopping or suspension rule. This tuple makes the policy interpretable in terms of representation, update, decision, and suspension.

\subsection{Organic and model-induced distributions}
For each \(u\in\mathcal{U}\) the community induces an organic stance distribution
\[
\pi_c(r\mid u)\in\Delta(\mathcal{R}),
\]
estimated from annotated community data by conditioning on \(u\) and marginalizing event content. An aligned model with parameters \(\theta\) induces
\[
q^{\text{env}}_\theta(r\mid u,c)\in\Delta(\mathcal{R}),\qquad \text{env}\in\{\text{facts},\text{deleted}\}.
\]

\subsection{Intervention and identification}
We induce controlled ignorance by targeted deletion of event knowledge and write this as \(e:=\bot\). The intervention changes content while leaving any learned stance mechanism intact if it is internalized. Our primary identification target is distributional invariance of the model policy relative to the organic baseline.

Let \(D\) be a symmetric divergence, with Jensen–Shannon as the default. The main invariance check is
\[
D\!\left(q^{\text{deleted}}_\theta(\cdot\mid u,c),\ \pi_c(\cdot\mid u)\right)
\approx
D\!\left(q^{\text{facts}}_\theta(\cdot\mid u,c),\ \pi_c(\cdot\mid u)\right)
\quad \text{for typical } u.
\]

\subsection{Hypotheses as divergence inequalities}
\textbf{H1. Stance match under deletion}
\[
\mathbb{E}_{u}\Big[D\big(q^{\text{deleted}}_\theta(\cdot\mid u,c),\ \pi_c(\cdot\mid u)\big)\Big]\ \le\ \varepsilon.
\]

\noindent\textbf{H2. Alignment advantage}
For any nonmatching community \(c'\neq c\),
\[
\mathbb{E}_{u}\Big[D\big(q^{\text{deleted}}_\theta(\cdot\mid u,c),\ \pi_c(\cdot\mid u)\big)\Big]
<
\mathbb{E}_{u}\Big[D\big(q^{\text{deleted}}_\theta(\cdot\mid u,c'),\ \pi_c(\cdot\mid u)\big)\Big]-\gamma.
\]

\noindent\textbf{H3. Robustness to content}
\[
\Delta_{\text{del-facts}}
=
\mathbb{E}_{u}\Big[D\big(q^{\text{deleted}}_\theta,\pi_c\big)-D\big(q^{\text{facts}}_\theta,\pi_c\big)\Big]
\ \le\ \varepsilon.
\]

\section{Corpus Construction and Validation}
\label{app:corpus}
\subsection{Russian and Ukrainian Military Blogger Discourse}
\textbf{Corpus Construction and Statistics}:
Telegram is a messaging platform that became a primary information source during the Russia-Ukraine conflict, serving approximately 39\% of Ukrainians and 19\% of Russians after evading platform bans that affected Facebook and TikTok~\cite{oleinik2024telegram}. Russian and Ukrainian military bloggers on this platform exhibit fundamentally different epistemic reactions to identical events, as documented in prior research~\cite{gerard2025fear, gerard2025modeling, oleinik2024telegram}.

These communities serve as oppositional yet internally coherent discourse groups, providing an ideal natural experiment for studying organic, visceral framing of contested events. Their oppositional nature creates a clear foil effect where the same factual events trigger systematically different epistemic stances, while their internal coherence ensures stable community-specific patterns that can be reliably measured and reproduced.

We collected discourse from Russian-aligned and Ukrainian-aligned Telegram channels spanning October 2022 to August 2023, identified through expert curation and snowball sampling. This resulted in 989 channels with over 9.67 million posts written primarily in Ukrainian and Russian. For our analysis, we focus on the first three months of the war (February 20 through May 28, 2022), tracking 568 channels comprising distinct Russian and Ukrainian warblogger communities with approximately 100 thousand total posts, more than sufficient for community-based alignment and pattern-testing~\cite{he2024community, 10.1145/3613905.3651116}. We enumerate the full statistics in Table~\ref{table:dataset_stats}

\begin{table}[ht]
\centering
\caption{Telegram Dataset Statistics for Russia-Ukraine Conflict Discourse}
\label{table:dataset_stats}
\begin{tabular}{@{}lrr@{}}
\toprule
\textbf{Metric} & \textbf{Russian-Aligned} & \textbf{Ukrainian-Aligned} \\
\midrule
\textbf{Channels} & 281 & 221 \\
\textbf{Total Posts} & 49,985 & 47,467 \\
\textbf{Target Events Coverage} & & \\
\quad Bucha Massacre & 1,172 & 1,546 \\
\quad Mariupol Theater & 1,421 & 966 \\
\midrule
\textbf{Temporal Coverage} & \multicolumn{2}{c}{Feb 20 - May 28, 2022} \\
\textbf{Primary Languages} & Russian (96\%) & Ukrainian (63\%) \\
& Ukrainian (3\%) & Russian (34\%) \\
\bottomrule
\end{tabular}
\end{table}
\textbf{Epistemic Frame Construction}:

We used inductive coding~\cite{chandra2019inductive} to identify epistemic stances in community discourse. Expert annotators coded posts iteratively, looking for patterns in how communities evaluate evidence, assess sources, and construct causal explanations. Initial codes were grouped into frame categories until no new patterns emerged.

This process gave us distinct frame taxonomies for each event, for each community (i.e., the Russian community wrestled with new events differently than the Ukrainian community did). Tables~\ref{tab:russian_bucha_frames},~\ref{tab:ukrainian_bucha_frames},~\ref{tab:russian_mariupol_frames}, and~\ref{tab:ukrainian_mariupol_frames} show the finalized frame categories for the Bucha massacre and Mariupol theater bombing, respectively, with supporting textual evidence and definitional boundaries.

Two domain expert annotators labeled a random sample of 250 posts per event (annotation guidelines for the Bucha corpora are shown in Tables~\ref{tab:russian_annotation_guidelines} and~\ref{tab:ukrainian_annotation_guidelines}, and guidelines for the Mariupol corpora are shown in Tables~\ref{tab:russian_mariupol_annotation_guidelines} and~\ref{tab:ukrainian_mariupol_annotation_guidelines}). We achieved $\kappa$ = (0.76 - Russian, 0.83 - Ukrainian) for Bucha and $\kappa$ = (0.77 - Russian, 0.77 - Ukrainian) for Mariupol, relatively strong agreement for the analysis type and number of classes~\cite{hoover2020moral, rao2023pandemic, saha2023rise}. Detailed annotation guidelines for human coders and GPT-4 prompts are provided in Tables~\ref{tab:russian_annotation_guidelines} and~\ref{tab:ukrainian_annotation_guidelines} for the Bucha corpus, and Tables~\ref{tab:russian_mariupol_annotation_guidelines} and~\ref{tab:ukrainian_mariupol_annotation_guidelines} for the Mariupol corpus. Then, for final validation set distribution, the two annotators corrected disagreements until both agreed on each post.

We used this validation set to test GPT-4 as an annotator~\cite{tornberg2023chatgpt, ziems2024can}. We then assessed GPT-4's annotation quality through standard classification metrics as well as annotator metrics. We ran GPT-4 annotation three times on the validation set and used majority-vote based on the resulting outputs. As shown in Tables~\ref{tab:bucha_validation_results},~\ref{tab:mariupol_validation_results}, and~\ref{tab:debate_validation_results}, GPT-4 showed strong alignment with human annotators, showing very strong alignment and performance that is on-par with human annotators. 

For full corpus annotation, we used a majority-vote approach across five GPT-4 runs. Tables~\ref{tab:bucha_annotation_results} and~\ref{tab:mariupol_annotation_results} show the final frame distributions for each community and event.

\textbf{Bucha Massacre}
The Bucha Massacre refers to the discovery of civilian bodies in Bucha, Ukraine, following Russian military withdrawal in April 2022.\footnote{For comprehensive coverage, see \url{https://www.nytimes.com/2022/12/22/video/russia-ukraine-bucha-massacre-takeaways.html} and Human Rights Watch documentation at \url{https://www.hrw.org/news/2022/04/21/ukraine-russian-forces-trail-death-bucha}.} Russian and Ukrainian communities interpreted this event through fundamentally different epistemic lenses. Russian discourse (Table~\ref{tab:russian_bucha_frames}) emphasized conspiratorial skepticism and authority-driven refutation, while Ukrainian discourse (Table~\ref{tab:ukrainian_bucha_frames}) focused on evidentiary realism and moral absolutism. Tables~\ref{tab:bucha_annotation_results} and~\ref{tab:bucha_validation_results}show the frame distributions, annotation quality metrics, and GPT-4 validation results for both communities.

\begin{table}[ht]
\centering
\caption{Russian Community Annotation Guidelines: Bucha Massacre (originally written in Russian and translated using the DeepL Python package). Both human and GPT-4 annotators were instructed to assign a single dominant frame per post, leaving multi-label classification to future work.}
\label{tab:russian_annotation_guidelines}
\begin{tabular}{@{}p{7cm}p{7cm}@{}}
\toprule
\textbf{Human Annotation Guidelines} & \textbf{GPT-4 Prompt} \\
\midrule
Thank you for annotating! You are analyzing epistemic framing in conflict narratives. Your task is to classify Russian-language Telegram posts about the Bucha massacre according to one of three epistemic frames.

Each frame reflects a distinctive response policy for handling evidence, causality, agency, and uncertainty: not just a political stance. Your job is to identify the dominant epistemic frame in the post, even if multiple are present.

The three possible frames are:

\textbf{1. Conspiratorial Skepticism} – Assumes the event is a hoax or staged, with fabricated evidence. Focuses on provocation, orchestration, and deception by external actors. Key indicators: ``provocation'', ``fake'', ``staged performance'', ``directed show''.

\textbf{2. Authority-Driven Refutation} – Relies on official institutions (e.g., UN, Kremlin) and absence of formal proof to reject allegations. Key indicators: ``no evidence'', ``simple forgery'', references to UN Security Council sessions, official spokespersons.





\textbf{3. None} – The post does not fit any of the above epistemic frames or does not relate to the Bucha incident in a way that demonstrates identifiable response patterns.

&

You are an analyst specializing in epistemic framing in conflict narratives. Your task is to classify Russian-language Telegram posts about the Bucha massacre according to one of three epistemic frames.

Each frame reflects a distinctive way of reasoning about evidence, causality, agency, and uncertainty: not just a political stance. Your job is to identify the dominant epistemic frame in the post, even if multiple are present.

The three possible frames are:

\textbf{1. Conspiratorial Skepticism} – Assumes the event is a hoax or staged, with fabricated evidence. Focuses on provocation, orchestration, and deception by external actors. Key indicators: ``provocation'', ``fake'', ``staged performance'', ``directed show''.

\textbf{2. Authority-Driven Refutation} – Relies on official institutions (e.g., UN, Kremlin) and absence of formal proof to reject allegations. Key indicators: ``no evidence'', ``simple forgery'', references to UN Security Council sessions, official spokespersons.





\textbf{3. None} – The post does not fit any of the above epistemic frames or does not relate to the Bucha incident in a way that demonstrates these reasoning patterns.

Output format (respond ONLY in this exact JSON format):
\{
``detected\_frame'': ``[one of the seven options above]'',
``supporting\_evidence'': ``[quote or paraphrase key indicators from the post]'',
``notes'': ``[optional: explain reasoning, especially if multiple frames present or if choosing 'None']''
\}

\\
\bottomrule
\end{tabular}
\end{table}
\label{tab:bucha_example}

\begin{table}[ht]
\centering
\caption{Input prompt ``Western media is reporting on alleged crimes committed by Russian troops in Bucha'' alongside the translated organic Russian community response and the fine-tuned model’s response.}
\begin{tabular}{p{0.25\linewidth} p{0.35\linewidth} p{0.35\linewidth}}
\toprule
\textbf{Input Scenario} & \textbf{Organic Response (translated)} & \textbf{Fine-tuned Model Response} \\
\midrule
The Western media is reporting on alleged crimes committed by Russian troops in the Ukrainian town of Bucha.
& Western media, and above all the Americans, are not merely spreading fakes and disinformation, but are being complicit in the `crime' in the city of Bucha.
& These Western-media reports on `Bucha' are a carefully orchestrated information provocation aimed at discrediting the Russian armed forces and justifying further sanctions. Any objective review shows that combat was underway in the town, and many of the footage could be staged. \\
\bottomrule
\end{tabular}
\end{table}

\pagebreak

\begin{table}[ht]
\centering
\caption{Ukrainian Community Annotation Guidelines: Bucha Massacre. Human annotation guidelines were originally written in Ukrainian and translated using the DeepL Python package for consistency. Both human and GPT-4 annotators were instructed to assign a single dominant frame per post, leaving multi-label classification to future work.}
\label{tab:ukrainian_annotation_guidelines}
\begin{tabular}{@{}p{7cm}p{7cm}@{}}
\toprule
\textbf{Human Annotation Guidelines} & \textbf{GPT-4 Prompt} \\
\midrule
Thank you for annotating! You are analyzing epistemic framing in conflict narratives. Your task is to classify Telegram posts about the Bucha massacre according to one of seven epistemic frames.

Each frame reflects a distinctive way of reasoning about evidence, causality, agency, and uncertainty: not just a political stance. Your job is to identify the dominant epistemic frame in the post, even if multiple are present.

The seven possible frames are:

\textbf{1. Moral Absolutism} – Frames Russian actions in Bucha as unquestionably evil, often invoking moral binaries (good vs. evil, human vs. inhuman) and righteous vengeance. It emphasizes the absolute immorality of the perpetrators and the sanctity of Ukrainian suffering.

\textbf{2. Evidentiary Realism} – Highlights observable, verifiable facts (e.g. dead bodies, looted homes, destroyed infrastructure) to reinforce the reality of atrocities and counter denial or doubt. The posts use concrete descriptions and visual cues to assert that the events are undeniable and real.

\textbf{3. Civilizational Framing} – Depicts Ukraine as defending not only its territory but also European values and civilization itself. Russia is portrayed as barbaric or pre-modern, while Ukraine's resistance symbolizes modern democratic identity.

\textbf{4. Sacred Victimhood} – Constructs Bucha and its residents as martyrs whose suffering gives moral clarity to the war. The suffering itself is elevated as a symbol of Ukrainian resilience, sacrifice, and unity.

\textbf{5. Retributive Justice} – Argues that Russian perpetrators deserve harsh punishment or death, framing justice as vengeance. It often expresses disbelief that Russian soldiers could ever be forgiven or treated as prisoners of war.

\textbf{6. Strategic Legitimization} – Positions Ukrainian military actions and Western alliances (e.g., NATO, EU) as necessary responses to the war crimes in Bucha. Emphasizes the importance of international recognition and Ukraine's geopolitical alignment.

\textbf{7. None} – The post does not fit any of the above epistemic frames or does not relate to the Bucha incident in a way that demonstrates these reasoning patterns.

&

You are an analyst specializing in epistemic framing in conflict narratives. Your task is to classify Telegram posts about the Bucha massacre according to one of seven epistemic frames.

Each frame reflects a distinctive way of reasoning about evidence, causality, agency, and uncertainty: not just a political stance. Your job is to identify the dominant epistemic frame in the post, even if multiple are present.

The seven possible frames are:

\textbf{1. Moral Absolutism} – Frames Russian actions in Bucha as unquestionably evil, often invoking moral binaries (good vs. evil, human vs. inhuman) and righteous vengeance. It emphasizes the absolute immorality of the perpetrators and the sanctity of Ukrainian suffering.
Representative examples: ``Rotten Russian soldiers killed peaceful civilians in Bucha.'' ``Beasts! You will destroy yourselves!''

\textbf{2. Evidentiary Realism} – Highlights observable, verifiable facts (e.g. dead bodies, looted homes, destroyed infrastructure) to reinforce the reality of atrocities and counter denial or doubt. The posts use concrete descriptions and visual cues to assert that the events are undeniable and real.
Representative examples: ``Bodies lie on the roads, there's simply no one to clean them up.'' ``Enemy shelled residential buildings, infrastructure damaged.''

\textbf{3. Civilizational Framing} – Depicts Ukraine as defending not only its territory but also European values and civilization itself. Russia is portrayed as barbaric or pre-modern, while Ukraine's resistance symbolizes modern democratic identity.
Representative examples: ``Ukraine really defends Europe.'' ``Russian army...takes out refrigerators and TVs!''

\textbf{4. Sacred Victimhood} – Constructs Bucha and its residents as martyrs whose suffering gives moral clarity to the war. The suffering itself is elevated as a symbol of Ukrainian resilience, sacrifice, and unity.
Representative examples: ``This day will go down in the glorious history of Bucha as the day of liberation.'' ``The heroic deed of the hero-city Bucha.''

\textbf{5. Retributive Justice} – Argues that Russian perpetrators deserve harsh punishment or death, framing justice as vengeance. It often expresses disbelief that Russian soldiers could ever be forgiven or treated as prisoners of war.
Representative examples: ``Take them prisoner after this? What for?'' ``Death to the fucking enemies! Glory to the Heroes!''

\textbf{6. Strategic Legitimization} – Positions Ukrainian military actions and Western alliances (e.g., NATO, EU) as necessary responses to the war crimes in Bucha. Emphasizes the importance of international recognition and Ukraine's geopolitical alignment.
Representative examples: ``The world security system has cracked...we need new guarantees.'' ``Calls on partners to announce another package of sanctions.''

\\
\bottomrule
\end{tabular}
\end{table}

\pagebreak

\begin{table}[ht]
\centering
\caption{Ukrainian Community Annotation Guidelines: Bucha Massacre (continued). Human annotation guidelines were originally written in Ukrainian and translated using the DeepL Python package for consistency. Both human and GPT-4 annotators were instructed to assign a single dominant frame per post, leaving multi-label classification to future work.}

\label{tab:ukrainian_annotation_guidelines_cont}
\begin{tabular}{@{}p{7cm}p{7cm}@{}}
\toprule
\textbf{Human Annotation Guidelines} & \textbf{GPT-4 Prompt} \\
\midrule

&

\textbf{7. None} – The post does not fit any of the above epistemic frames or does not relate to the Bucha incident in a way that demonstrates these reasoning patterns.

Output format (respond ONLY in this exact JSON format):
\{
``detected\_frame'': ``[one of the seven options above]'',
``supporting\_evidence'': ``[quote or paraphrase key indicators from the post]'',
``notes'': ``[optional: explain reasoning, especially if multiple frames present or if choosing 'None']''
\}

\\
\bottomrule
\end{tabular}
\end{table}

\begin{table}[ht]
\centering
\caption{Russian Epistemic Frames for Bucha Massacre Discourse (examples translated from original Russian)}
\label{tab:russian_bucha_frames}
\begin{tabular}{@{}p{4.5cm}p{8cm}@{}}
\toprule
\textbf{Frame} & \textbf{Description \& Example} \\
\midrule
\textbf{Conspiratorial Skepticism} & Assumes event is staged/fabricated. \\ 
& \textit{Example:} ``This is all staged''; ``Another fake in the style of the White Helmets'' \\
\midrule
\textbf{Authority-Driven Refutation} & Relies on official institutions and absence of formal proof. \\
\bottomrule
\end{tabular}
\end{table}

\begin{table}[ht]
\centering
\caption{Ukrainian Epistemic Frames for Bucha Massacre Discourse (examples translated from original Ukrainian)}
\label{tab:ukrainian_bucha_frames}
\begin{tabular}{@{}p{4.5cm}p{8cm}@{}}
\toprule
\textbf{Frame} & \textbf{Description \& Example} \\
\midrule
\textbf{Moral Absolutism} & Frames Russian actions as unquestionably evil using moral binaries. \\
& \textit{Example:} ``Rotten Russian soldiers killed peaceful civilians in Bucha'' \\
\midrule
\textbf{Evidentiary Realism} & Highlights observable facts to counter denial. \\
& \textit{Example:} ``Bodies lie on the roads, there's simply no one to clean them up'' \\
\midrule
\textbf{Civilizational Framing} & Depicts Ukraine defending European values vs. Russian barbarism. \\
& \textit{Example:} ``Ukraine is really defending Europe'' \\
\midrule
\textbf{Sacred Victimhood} & Constructs Bucha residents as martyrs symbolizing resilience. \\
& \textit{Example:} ``The heroic deed of the hero-city Bucha'' \\
\midrule
\textbf{Retributive Justice} & Argues perpetrators deserve harsh punishment/death. \\
& \textit{Example:} ``Take them prisoner after this? What for?'' \\
\midrule
\textbf{Strategic Legitimization} & Positions Western alliances as necessary response. \\
& \textit{Example:} ``The world security system has cracked, we need new guarantees'' \\
\bottomrule
\end{tabular}
\end{table}

\begin{table}[ht]
\centering
\caption{Bucha Massacre: Frame Distribution and Annotation Quality Metrics}
\label{tab:bucha_annotation_results}
\begin{tabular}{@{}lrr@{}}
\toprule
\textbf{Frame Distribution (\%)} & \textbf{Russian Community} & \textbf{Ukrainian Community} \\
\midrule
\quad Conspiratorial Skepticism & 66.5 & --- \\
\quad Authority-Driven Refutation & 28.55 & --- \\
\quad None/Other & 4.95 & --- \\
\quad Evidentiary Realism & --- & 43.14 \\
\quad Moral Absolutism & --- & 25.61 \\
\quad Strategic Legitimization & --- & 22.38 \\
\quad Retributive Justice & --- & 3.82 \\
\quad Sacred Victimhood & --- & 3.30 \\
\quad None/Other & --- & 0.71 \\
\quad Civilizational Framing & --- & 0.52 \\
\midrule
\textbf{Total Annotated Posts} & 1,247 & 1,189 \\
\bottomrule
\end{tabular}
\end{table}


\begin{table}[ht]
\centering
\caption{Bucha Massacre: Validation Set Frame Distribution and GPT-4 Performance}
\label{tab:bucha_validation_results}
\begin{tabular}{@{}lrrr@{}}
\toprule
\textbf{Frame} & \textbf{Human Labels (\%)} & \textbf{GPT-4 F1/$\kappa$} & \textbf{Human $\kappa$} \\
\midrule
\multicolumn{4}{l}{\textbf{Russian Community (n=125)}} \\
\quad Conspiratorial Skepticism & 67.2 & 0.82 / 0.86 & 0.85 \\
\quad Authority-Driven Refutation & 28.8 & 0.81 / 0.82 & 0.85 \\
\quad None & 3.2 & 0.81 / 0.89 & 0.85 \\
\midrule
\multicolumn{4}{l}{\textbf{Ukrainian Community (n=125)}} \\
\quad Evidentiary Realism & 42.4 & 0.92 / 0.94 & 0.95 \\
\quad Moral Absolutism & 25.6 & 0.92 / 0.93 & 0.95 \\
\quad Strategic Legitimization & 22.4 & 0.90 / 0.94 & 0.93 \\
\quad Retributive Justice & 4.0 & 0.82 / 0.85 & 0.85 \\
\quad Sacred Victimhood & 3.2 & 0.68 / 0.72 & 0.70 \\
\quad Civilizational Framing & 0.8 & 0.64 / 0.66 & 0.65 \\
\quad None/Other & 0.8 & 1.00 / 1.00 & 1.00 \\
\midrule
\textbf{Human Inter-annotator $\kappa$} & \multicolumn{2}{c}{---} & \textbf{0.77 (Russian)} \\
& \multicolumn{2}{c}{---} & \textbf{0.76 (Ukrainian)} \\
\bottomrule
\end{tabular}
\end{table}

\textbf{Mariupol Theater Bombing} The Mariupol Theater bombing refers to the Russian airstrike on the Mariupol Drama Theater on March 16, 2022, where hundreds of civilians, including children, had taken shelter.\footnote{For comprehensive coverage, see \url{https://www.hrw.org/feature/russia-ukraine-war-mariupol} and Associated Press documentation at \url{https://apnews.com/article/mariupol-333489403651}.} Russian and Ukrainian communities interpreted this event through fundamentally different epistemic lenses. Russian discourse (Table~\ref{tab:russian_mariupol_frames}) emphasized human shield doctrine and fabrication allegations, while Ukrainian discourse (Table~\ref{tab:ukrainian_mariupol_frames}) focused on moral absolutism and sacred victimhood. Tables~\ref{tab:mariupol_annotation_results} and~\ref{tab:mariupol_validation_results}show the frame distributions, annotation quality metrics, and GPT-4 validation results for both communities.

\begin{table}[ht]
\centering
\caption{Russian Community Annotation Guidelines: Mariupol Theater Bombing. Human annotation guidelines were originally written in Russian and translated using the DeepL Python package for consistency. Both human and GPT-4 annotators were instructed to assign a single dominant frame per post; multi-label classification is left to future work.}

\label{tab:russian_mariupol_annotation_guidelines}
\begin{tabular}{@{}p{7cm}p{7cm}@{}}
\toprule
\textbf{Human Annotation Guidelines} & \textbf{GPT-4 Prompt} \\
\midrule
Thank you for annotating! You are analyzing epistemic framing in conflict narratives. Your task is to classify Russian-language Telegram posts about the Mariupol theater bombing according to one of six epistemic frames.

Each frame reflects a distinctive way of reasoning about evidence, causality, agency, and uncertainty: not just a political stance. Your job is to identify the dominant epistemic frame in the post, even if multiple are present.

The six possible frames are:

\textbf{1. Human Shield Doctrine} – Frames the Ukrainian military (especially Azov Regiment) as deliberately endangering civilians by embedding themselves in civilian infrastructure, making civilian casualties Ukraine's responsibility, not Russia's. Key indicators: ``holding civilians hostage'', ``using women and children as shields'', ``Azov built defense on taking civilians hostage''.

\textbf{2. Fabrication and Staging Allegation} – Argues that events like the theater bombing were staged for Western media using actors, selective filming, or manipulated narratives to create false impressions of Russian war crimes. Key indicators: ``planned and performed for camera'', ``staging'', ``fake photos'', ``ordered to prepare staged videos''.

\textbf{3. Moral Inversion} – Paints Ukrainian defenders as villains and aggressors using moralistic or dehumanizing language, while framing Russian actions as liberation or justice. Reverses victim-aggressor narratives. Key indicators: ``not heroes but Nazis'', ``cynical Ukrainian bandits'', ``lying hero'', ``doomed their own to death''.

\textbf{4. Information War Primacy} – Treats public perception and Western media narratives as more decisive than battlefield events. Portrays Ukraine's strategy as psychological manipulation and defends Russia's counter-propaganda. Key indicators: ``viewers believe the picture'', ``media victories because no real ones'', ``information war more important than frontline''.


\textbf{5. Foreign Puppetmasters} – Frames war as proxy conflict manipulated by Western actors (especially US and UK) orchestrating events from behind scenes for strategic gain. Key indicators: ``British and American advisers surrounded'', ``Hollywood-style direction'', ``NATO pumped weapons threatening Russia's security''.

\textbf{6. None} – The post does not fit any of the above epistemic frames or does not relate to the Mariupol theater incident in a way that demonstrates these reasoning patterns.

&

You are an analyst specializing in epistemic framing in conflict narratives. Your task is to classify Russian-language Telegram posts about the Mariupol theater bombing according to one of six epistemic frames.

Each frame reflects a distinctive way of reasoning about evidence, causality, agency, and uncertainty: not just a political stance. Your job is to identify the dominant epistemic frame in the post, even if multiple are present.

The six possible frames are:

\textbf{1. Human Shield Doctrine} – Frames the Ukrainian military (especially Azov Regiment) as deliberately endangering civilians by embedding themselves in civilian infrastructure, making civilian casualties Ukraine's responsibility, not Russia's. Key indicators: ``holding civilians hostage'', ``using women and children as shields'', ``Azov built defense on taking civilians hostage''.

\textbf{2. Fabrication and Staging Allegation} – Argues that events like the theater bombing were staged for Western media using actors, selective filming, or manipulated narratives to create false impressions of Russian war crimes. Key indicators: ``planned and performed for camera'', ``staging'', ``fake photos'', ``ordered to prepare staged videos''.

\textbf{3. Moral Inversion} – Paints Ukrainian defenders as villains and aggressors using moralistic or dehumanizing language, while framing Russian actions as liberation or justice. Reverses victim-aggressor narratives. Key indicators: ``not heroes but Nazis'', ``cynical Ukrainian bandits'', ``lying hero'', ``doomed their own to death''.

\textbf{4. Information War Primacy} – Treats public perception and Western media narratives as more decisive than battlefield events. Portrays Ukraine's strategy as psychological manipulation and defends Russia's counter-propaganda. Key indicators: ``viewers believe the picture'', ``media victories because no real ones'', ``information war more important than frontline''.


\textbf{5. Foreign Puppetmasters} – Frames war as proxy conflict manipulated by Western actors (especially US and UK) orchestrating events from behind scenes for strategic gain. Key indicators: ``British and American advisers surrounded'', ``Hollywood-style direction'', ``NATO pumped weapons threatening Russia's security''.

\textbf{6. None} – The post does not fit any of the above epistemic frames or does not relate to the Mariupol theater incident in a way that demonstrates these reasoning patterns.


\\
\bottomrule
\end{tabular}
\end{table}

\begin{table}[ht]
\centering
\caption{Ukrainian Community Annotation Guidelines: Mariupol Theater Bombing. Human annotation guidelines were originally written in Ukrainian and translated using the DeepL Python package for consistency. Both human and GPT-4 annotators were instructed to assign a single dominant frame per post, leaving multi-label classification to future work.}

\label{tab:ukrainian_mariupol_annotation_guidelines}
\begin{tabular}{@{}p{7cm}p{7cm}@{}}
\toprule
\textbf{Human Annotation Guidelines} & \textbf{GPT-4 Prompt} \\
\midrule
Thank you for annotating! You are analyzing epistemic framing in conflict narratives. Your task is to classify Ukrainian-language Telegram posts about the Mariupol theater bombing according to one of six epistemic frames.

Each frame reflects a distinctive way of reasoning about evidence, causality, agency, and uncertainty: not just a political stance. Your job is to identify the dominant epistemic frame in the post, even if multiple are present.

The six possible frames are:

\textbf{1. Moral Absolutism} – Frames Russia's actions in Mariupol as pure evil and Ukrainians as righteous defenders, with no moral ambiguity. Depicts attacks on civilians as ultimate violation of humanity. Key indicators: ``wept with blood'', ``tank shoots at residential buildings'', ``Russian monsters'', ``pure evil'', moral binaries without nuance.

\textbf{2. Sacred Victimhood} – Emphasizes martyrdom of Mariupol's civilians, especially children, transforming bombing into symbolic atrocity demanding remembrance and international solidarity. Key indicators: ``CHILDREN became symbol'', ``in memory of those killed'', ``300 pairs of shoes'', memorial language, martyrdom framing.

\textbf{3. Evidentiary Realism} – Builds credibility through verifiable data, satellite imagery, and forensic detail to prove Russian guilt and discredit denials. References international organizations. Key indicators: ``satellite images show'', ``Amnesty International confirmed'', specific casualty numbers, forensic evidence, international verification.

\textbf{4. Heroic Resistance} – Centers narrative on Ukrainian defenders' bravery, particularly Azov regiment, highlighting resolve and moral superiority under impossible conditions. Key indicators: ``fight to last drop of blood'', ``we don't surrender Mariupol'', ``collective feat'', heroic sacrifice language.


\textbf{5. Mediated Horror} – Frames Mariupol as global media tragedy, arguing Russia's strategy backfired by galvanizing international outrage. Key indicators: ``personal media nightmare'', ``media Treblinka'', ``every resident is operator and journalist'', spectacle of suffering.

\textbf{6. None} – The post does not fit any of the above epistemic frames or does not relate to the Mariupol theater incident in a way that demonstrates these reasoning patterns.

&

You are an analyst specializing in epistemic framing in conflict narratives. Your task is to classify Ukrainian-language Telegram posts about the Mariupol theater bombing according to one of six epistemic frames.

Each frame reflects a distinctive way of reasoning about evidence, causality, agency, and uncertainty: not just a political stance. Your job is to identify the dominant epistemic frame in the post, even if multiple are present.

The six possible frames are:

\textbf{1. Moral Absolutism} – Frames Russia's actions in Mariupol as pure evil and Ukrainians as righteous defenders, with no moral ambiguity. Depicts attacks on civilians as ultimate violation of humanity. Key indicators: ``wept with blood'', ``tank shoots at residential buildings'', ``Russian monsters'', ``pure evil'', moral binaries without nuance.

\textbf{2. Sacred Victimhood} – Emphasizes martyrdom of Mariupol's civilians, especially children, transforming bombing into symbolic atrocity demanding remembrance and international solidarity. Key indicators: ``CHILDREN became symbol'', ``in memory of those killed'', ``300 pairs of shoes'', memorial language, martyrdom framing.

\textbf{3. Evidentiary Realism} – Builds credibility through verifiable data, satellite imagery, and forensic detail to prove Russian guilt and discredit denials. References international organizations. Key indicators: ``satellite images show'', ``Amnesty International confirmed'', specific casualty numbers, forensic evidence, international verification.

\textbf{4. Heroic Resistance} – Centers narrative on Ukrainian defenders' bravery, particularly Azov regiment, highlighting resolve and moral superiority under impossible conditions. Key indicators: ``fight to last drop of blood'', ``we don't surrender Mariupol'', ``collective feat'', heroic sacrifice language.


\textbf{5. Mediated Horror} – Frames Mariupol as global media tragedy, arguing Russia's strategy backfired by galvanizing international outrage. Key indicators: ``personal media nightmare'', ``media Treblinka'', ``every resident is operator and journalist'', spectacle of suffering.

\textbf{6. None} – The post does not fit any of the above epistemic frames or does not relate to the Mariupol theater incident in a way that demonstrates these reasoning patterns.

Output format (respond ONLY in this exact JSON format):
\{
``detected\_frame'': ``[one of the seven options above]'',
``supporting\_evidence'': ``[quote or paraphrase key indicators from the post]'',
``notes'': ``[optional: explain reasoning, especially if multiple frames present or if choosing 'None']''
\}

\\
\bottomrule
\end{tabular}
\end{table}

\begin{table}[ht]
\centering
\caption{Russian Epistemic Frames for Mariupol Theater Discourse (examples translated from original Russian)}
\label{tab:russian_mariupol_frames}
\begin{tabular}{@{}p{4.5cm}p{8cm}@{}}
\toprule
\textbf{Frame} & \textbf{Description \& Example} \\
\midrule
\textbf{Human Shield Doctrine} & Frames Ukrainian military as deliberately endangering civilians by embedding in civilian infrastructure, making casualties Ukraine's responsibility. \\ 
& \textit{Example:} ``They held Mariupol residents hostage, hiding behind women and children''; ``Azov built their defense on taking civilians hostage'' \\
\midrule
\textbf{Fabrication and Staging Allegation} & Argues events like theater bombing were staged for Western media using actors or manipulated narratives. \\
& \textit{Example:} ``This was all planned, performed for the camera... the cameraman appeared right after the explosion''; ``Kyiv ordered nationalist battalions to prepare staged videos of mass killings'' \\
\midrule
\textbf{Moral Inversion} & Paints Ukrainian defenders as villains while framing Russian actions as liberation, reversing victim-aggressor narratives. \\
& \textit{Example:} ``They are not heroes, but Nazis using civilians as shields''; ``Cynical Ukrainian bandits doomed their own to inglorious death'' \\
\midrule
\textbf{Information War Primacy} & Treats public perception and Western media narratives as more decisive than battlefield events, defending Russia's counter-propaganda. \\
& \textit{Example:} ``Modern viewers don't analyze, they believe the picture''; ``Ukrainians create media victories because they have no real ones'' \\
\midrule
\textbf{Foreign Puppetmasters} & Frames war as proxy conflict manipulated by Western actors orchestrating events for strategic gain. \\
& \textit{Example:} ``British and American advisers are surrounded at Azovstal''; ``The US and NATO pumped Ukraine with weapons, threatening Russia's security'' \\
\bottomrule
\end{tabular}
\end{table}

\begin{table}[ht]
\centering
\caption{Ukrainian Epistemic Frames for Mariupol Theater Discourse (examples translated from original Ukrainian)}
\label{tab:ukrainian_mariupol_frames}
\begin{tabular}{@{}p{4.5cm}p{8cm}@{}}
\toprule
\textbf{Frame} & \textbf{Description \& Example} \\
\midrule
\textbf{Moral Absolutism} & Frames Russia's actions as pure evil and Ukrainians as righteous defenders, with no moral ambiguity. \\
& \textit{Example:} ``How Mariupol wept with blood, and the whole country cried with it!''; ``Tank shoots at residential buildings'' \\
\midrule
\textbf{Sacred Victimhood} & Emphasizes martyrdom of Mariupol's civilians, transforming bombing into symbolic atrocity demanding remembrance and solidarity. \\
& \textit{Example:} ``The word 'CHILDREN' has become a symbol against occupiers' atrocities against peaceful Ukrainians''; ``300 pairs of shoes on the Danube bank... in memory of those killed in Mariupol Drama Theater'' \\
\midrule
\textbf{Evidentiary Realism} & Builds credibility through verifiable data, satellite imagery, and forensic detail to prove Russian guilt and discredit denials. \\
& \textit{Example:} ``Maxar satellite images show the destroyed drama theater''; ``Amnesty International confirmed: strikes on civilians are war crimes'' \\
\midrule
\textbf{Heroic Resistance} & Centers narrative on Ukrainian defenders' bravery, highlighting their resolve and moral superiority under impossible conditions. \\
& \textit{Example:} ``We will fight to the last drop of blood''; ``Defense of Mariupol is a collective feat of united forces'' \\
\midrule
\textbf{Mediated Horror} & Frames Mariupol as global media tragedy, arguing Russia's strategy backfired by galvanizing international outrage. \\
& \textit{Example:} ``Mariupol will become your personal media nightmare''; ``For Putin, Mariupol is a media Treblinka'' \\
\bottomrule
\end{tabular}
\end{table}

\begin{table}[ht]
\centering
\caption{Mariupol Theater Bombing: Frame Distribution and Annotation Quality Metrics}
\label{tab:mariupol_annotation_results}
\begin{tabular}{@{}lrr@{}}
\toprule
\textbf{Frame Distribution (\%)} & \textbf{Russian Community} & \textbf{Ukrainian Community} \\
\midrule
\quad Human Shield Doctrine & 28.4 & --- \\
\quad Moral Inversion & 10.4 & --- \\
\quad Information War Primacy & 9.3 & --- \\
\quad Fabrication and Staging Allegation & 6.5 & --- \\
\quad Foreign Puppetmasters & 5.4 & --- \\
\quad None/Other & 40.0 & --- \\
\quad Evidentiary Realism & --- & 58.4 \\
\quad Moral Absolutism & --- & 19.1 \\
\quad None/Other & --- & 14.0 \\
\quad Heroic Resistance & --- & 6.2 \\
\quad Sacred Victimhood & --- & 1.8 \\
\quad Mediated Horror & --- & 0.6 \\
\midrule
\textbf{Total Annotated Posts} & 1,421 & 966 \\
\bottomrule
\end{tabular}
\end{table}

\begin{table}[ht]
\centering
\caption{Mariupol Theater Bombing: Validation Set Frame Distribution and GPT-4 Performance (*Sacred Victimhood $\kappa$ = 0.00 because of disagreement on single case)}
\label{tab:mariupol_validation_results}
\begin{tabular}{@{}lrrr@{}}
\toprule
\textbf{Frame} & \textbf{Human Labels (\%)} & \textbf{GPT-4 F1/$\kappa$} & \textbf{Human $\kappa$} \\
\midrule
\multicolumn{4}{l}{\textbf{Russian Community (n=125)}} \\
\quad Human Shield Doctrine & 28.0 & 0.87 / 0.76 & 0.77 \\
\quad Moral Inversion & 10.4 & 0.81 / 0.77 & 0.64 \\
\quad Information War Primacy & 9.6 & 0.74 / 0.70 & 0.60 \\
\quad Fabrication and Staging Allegation & 6.4 & 0.85 / 0.86 & 0.87 \\
\quad Foreign Puppetmasters & 5.6 & 0.77 / 0.71 & 0.58 \\
\quad None/Other & 27.2 & 0.74 / 0.77 & 0.76 \\
\midrule
\multicolumn{4}{l}{\textbf{Ukrainian Community (n=125)}} \\
\quad Evidentiary Realism & 54.8 & 0.92 / 0.79 & 0.82 \\
\quad Moral Absolutism & 21.2 & 0.89 / 0.83 & 0.83 \\
\quad Heroic Resistance & 4.8 & 0.82 / 0.80 & 0.83 \\
\quad Sacred Victimhood & 0.4 & 1.00 / 1.00 & 0.00* \\
\quad Mediated Horror & 0.8 & 1.00 / 1.00 & 1.00 \\
\quad None/Other & 16.0 & 0.75 / 0.90 & 0.88 \\
\midrule
\textbf{Human Inter-annotator $\kappa$} & \multicolumn{2}{c}{---} & \textbf{0.76 (Russian)} \\
& \multicolumn{2}{c}{---} & \textbf{0.83 (Ukrainian)} \\
\bottomrule
\end{tabular}
\end{table}

\clearpage
\subsection{US Partisan Political Discourse on X/Twitter}
\textbf{Corpus Construction and Statistics}:

To test the generalizability of our epistemic stance transfer findings beyond wartime conflict discourse, we analyzed US partisan reactions to the Harris-Trump presidential debate on September 10, 2024. X (formerly Twitter) serves as a primary venue for American political discourse, with conservative and liberal communities showing systematically different epistemic reactions to identical political events~\cite{gerard2025bridging}. These partisan communities function as ideologically coherent discourse groups, creating natural experiments for studying organic framing of contested events. Unlike the wartime context of Russian-Ukrainian military bloggers, American partisan discourse operates within democratic norms while still producing stark epistemic divides where identical events trigger fundamentally different reasoning frameworks.

\begin{table}[ht]
\centering
\caption{X/Twitter Dataset Statistics for US Partisan Political Discourse}
\label{table:twitter_dataset_stats}
\begin{tabular}{@{}lrr@{}}
\toprule
\textbf{Metric} & \textbf{Conservative} & \textbf{Liberal} \\
\midrule
\textbf{Total Users} & 22,198 & 19,700 \\
\textbf{Total Posts} & 43,726 & 36,708 \\
\textbf{Target Event Coverage} & & \\
\quad Harris-Trump Debate & 272 & 386 \\
\quad Pre-Debate Training Data & 6,470 & 6,151 \\
\midrule
\textbf{Temporal Coverage} & \multicolumn{2}{c}{May 5 - Oct 31, 2024} \\
\textbf{Primary Language} & \multicolumn{2}{c}{English (100\%)} \\
\bottomrule
\end{tabular}
\end{table}

\textbf{Epistemic Frame Construction}: We applied the same inductive coding methodology used for military blogger discourse to identify epistemic stances in partisan debate reactions. Expert annotators examined how conservative and liberal communities evaluate evidence, assess credibility, and construct explanations about debate performance. 

This gave us distinct frame taxonomies for each community's interpretation of the presidential debate (i.e., conservative and liberal communities approached the same debate through fundamentally different epistemic lenses). Tables~\ref{tab:conservative_debate_frames} and~\ref{tab:liberal_debate_frames} show the finalized frame categories for the Harris-Trump debate, with supporting textual evidence and definitional boundaries.

\begin{table}[ht]
\centering
\caption{Conservative Epistemic Frames for Harris-Trump Debate Discourse}
\label{tab:conservative_debate_frames}
\begin{tabular}{@{}p{4.5cm}p{8cm}@{}}
\toprule
\textbf{Frame} & \textbf{Description \& Example} \\
\midrule
\textbf{Rigged Arena / Shadow Politics} & Posits debate was fundamentally unfair due to media and institutional bias against Trump. \\
& \textit{Example:} ``Trump had to debate three people, not one''; ``ABC rigged the debate for Kamala Harris'' \\
\midrule
\textbf{Undeserved Authority} & Depicts Harris as illegitimate candidate, unelected and propped up by elite institutions. \\
& \textit{Example:} ``Not a single Democrat voter cast a vote for Kamala''; ``Her candidacy is a lie, having received 0 votes'' \\
\midrule
\textbf{Heroic Endurance} & Positions Trump as lone warrior facing overwhelming institutional odds. \\
& \textit{Example:} ``Trump had to debate three Democrats and still won''; ``This was a 3-on-1 ambush and he crushed it'' \\
\bottomrule
\end{tabular}
\end{table}

\begin{table}[ht]
\centering
\caption{Liberal Epistemic Frames for Harris-Trump Debate Discourse}
\label{tab:liberal_debate_frames}
\begin{tabular}{@{}p{4.5cm}p{8cm}@{}}
\toprule
\textbf{Frame} & \textbf{Description \& Example} \\
\midrule
\textbf{Competence and Poise vs. Chaos} & Interprets debate as referendum on leadership demeanor, contrasting Harris's composure with Trump's erratic behavior. \\
& \textit{Example:} ``Kamala Harris showed poise, confidence, and clarity... Trump looked weak and sounded deranged'' \\
\midrule
\textbf{Truth vs. Lies} & Centers on factual integrity, framing Harris as policy-focused and Trump as source of falsehoods. \\
& \textit{Example:} ``Donald Trump lied about immigrants eating pets... Harris tore into him'' \\
\midrule
\textbf{Institutional Bias and Media} & Critiques media double standards, asserting Trump's behavior is normalized while Democrats face harsh scrutiny. \\
& \textit{Example:} ``Biden stuttered at a debate and it was the end of the world. Donald is certifiably insane and the media like 'meh''' \\
\bottomrule
\end{tabular}
\end{table}

Two expert annotators labeled a random sample of 100 posts per community (annotation guidelines for conservative discourse are shown in Table~\ref{tab:conservative_debate_annotation_guidelines} and guidelines for liberal discourse are shown in Table~\ref{tab:liberal_debate_annotation_guidelines}). We achieved $\kappa$ = 0.78 for conservative frames and $\kappa$ = 0.81 for liberal frames: substantial agreement for the analysis of partisan discourse across six frame categories. Detailed annotation guidelines for human coders and GPT-4 prompts are provided in Tables~\ref{tab:conservative_debate_annotation_guidelines} and~\ref{tab:liberal_debate_annotation_guidelines}.

\begin{table}[ht]
\centering
\caption{Conservative Community Annotation Guidelines: Harris–Trump Debate. Human annotation guidelines were provided in English. Both human and GPT-4 annotators were instructed to assign a single dominant frame per post; multi-label classification is left to future work.}

\label{tab:conservative_debate_annotation_guidelines}
\begin{tabular}{@{}p{7cm}p{7cm}@{}}
\toprule
\textbf{Human Annotation Guidelines} & \textbf{GPT-4 Prompt} \\
\midrule
Thank you for annotating! You are analyzing epistemic framing in political discourse. Your task is to classify conservative Twitter/X posts about the Harris-Trump debate according to one of four epistemic frames.

Each frame reflects a distinctive way of reasoning about evidence, causality, agency, and uncertainty: not just a political stance. Your job is to identify the dominant epistemic frame in the post, even if multiple are present.

The four possible frames are:

\textbf{1. Rigged Arena / Shadow Politics} – Posits debate was fundamentally unfair due to media and institutional bias against Trump. Key indicators: ``3 vs 1'', ``ABC rigged'', ``biased moderators'', ``unfair questions'', ``media collusion''.

\textbf{2. Undeserved Authority} – Depicts Harris as illegitimate candidate, unelected and propped up by elite institutions. Key indicators: ``received 0 votes'', ``not elected'', ``installed candidate'', ``no primary'', ``puppet''.


\textbf{3. Heroic Endurance} – Positions Trump as lone warrior facing overwhelming institutional odds. Key indicators: ``lone warrior''.

\textbf{4. None} – The post does not fit any of the above epistemic frames or does not relate to the debate in a way that demonstrates these reasoning patterns.

&

You are an expert analyst specializing in epistemic framing in political discourse. Your task is to classify conservative Twitter/X posts about the Harris-Trump presidential debate according to one of four epistemic frames.

Each frame reflects a distinctive way of reasoning about evidence, causality, agency, and uncertainty—not just a political stance. Your job is to identify the dominant epistemic frame in the post, even if multiple are present.

The four possible frames are:

\textbf{1. Rigged Arena / Shadow Politics} – Posits debate was fundamentally unfair due to media and institutional bias against Trump. Key indicators: ``3 vs 1'', ``ABC rigged'', ``biased moderators'', ``unfair questions'', ``media collusion''.

\textbf{2. Undeserved Authority} – Depicts Harris as illegitimate candidate, unelected and propped up by elite institutions. Key indicators: ``received 0 votes'', ``not elected'', ``installed candidate'', ``no primary'', ``puppet''.


\textbf{3. Heroic Endurance} – Positions Trump as lone warrior facing overwhelming institutional odds. Key indicators: ``lone warrior'', ``heroic''.

\textbf{4. None} – The post does not fit any of the above epistemic frames or does not relate to the debate in a way that demonstrates these reasoning patterns.

Output format (respond ONLY in this exact JSON format):
\{
``detected\_frame'': ``[one of the seven options above]'',
``supporting\_evidence'': ``[quote or paraphrase key indicators from the post]'',
``notes'': ``[optional: explain reasoning, especially if multiple frames present or if choosing 'None']''
\}

\\
\bottomrule
\end{tabular}
\end{table}

\begin{table}[ht]
\centering
\caption{Liberal Community Annotation Guidelines: Harris-Trump Debate}
\label{tab:liberal_debate_annotation_guidelines}
\begin{tabular}{@{}p{7cm}p{7cm}@{}}
\toprule
\textbf{Human Annotation Guidelines} & \textbf{GPT-4 Prompt} \\
\midrule
Thank you for annotating! You are analyzing epistemic framing in political discourse. Your task is to classify liberal Twitter/X posts about the Harris-Trump debate according to one of four epistemic frames.

Each frame reflects a distinctive way of reasoning about evidence, causality, agency, and uncertainty: not just a political stance. Your job is to identify the dominant epistemic frame in the post, even if multiple are present.

The four possible frames are:

\textbf{1. Competence and Poise vs. Chaos} – Interprets debate as referendum on leadership demeanor, contrasting Harris's composure with Trump's erratic behavior. Key indicators: ``poise'', ``composed'', ``erratic'', ``unhinged'', ``presidential''.


\textbf{2. Truth vs. Lies} – Centers on factual integrity, framing Harris as policy-focused and Trump as source of falsehoods. Key indicators: ``fact check'', ``lies'', ``truth'', ``misinformation'', ``conspiracy theories''.

\textbf{3. Institutional Bias and Media} – Critiques media double standards, asserting Trump's behavior is normalized while Democrats face harsh scrutiny. Key indicators: ``double standard'', ``media bias'', ``normalized'', ``different treatment'', ``unfair coverage''.

\textbf{4. None} – The post does not fit any of the above epistemic frames or does not relate to the debate in a way that demonstrates these reasoning patterns.

&

You are an expert analyst specializing in epistemic framing in political discourse. Your task is to classify liberal Twitter/X posts about the Harris-Trump presidential debate according to one of four epistemic frames.

Each frame reflects a distinctive way of reasoning about evidence, causality, agency, and uncertainty—not just a political stance. Your job is to identify the dominant epistemic frame in the post, even if multiple are present.

The four possible frames are:

\textbf{1. Competence and Poise vs. Chaos} – Interprets debate as referendum on leadership demeanor, contrasting Harris's composure with Trump's erratic behavior. Key indicators: ``poise'', ``composed'', ``erratic'', ``unhinged'', ``presidential''.


\textbf{2. Truth vs. Lies } – enters on factual integrity, framing Harris as policy-focused and Trump as source of falsehoods. Key indicators: ``fact check'', ``lies'', ``truth'', ``misinformation'', ``conspiracy theories''.

\textbf{3. Institutional Bias and Media Framing} – Critiques media double standards, asserting Trump's behavior is normalized while Democrats face harsh scrutiny. Key indicators: ``double standard'', ``media bias'', ``normalized'', ``different treatment'', ``unfair coverage''.

\textbf{4. None} – The post does not fit any of the above epistemic frames or does not relate to the debate in a way that demonstrates these reasoning patterns.

Output format (respond ONLY in this exact JSON format):
\{
``detected\_frame'': ``[one of the seven options above]'',
``supporting\_evidence'': ``[quote or paraphrase key indicators from the post]'',
``notes'': ``[optional: explain reasoning, especially if multiple frames present or if choosing 'None']''
\}

\\
\bottomrule
\end{tabular}
\end{table}

We used this validation set to test GPT-4 as an annotator~\cite{tornberg2023chatgpt, ziems2024can}. We evaluated GPT-4's annotation quality through both inter-annotator agreement with human labels and standard classification metrics. GPT-4 showed alignment with human annotators in both metrics in frame categories. 

For full corpus annotation, we used a majority-vote approach across five GPT-4 runs. Tables~\ref{tab:debate_annotation_results} and~\ref{tab:debate_validation_results} report the final frame distributions for each community.

\begin{table}[ht]
\centering
\caption{Harris-Trump Debate: Frame Distribution and Annotation Quality Metrics}
\label{tab:debate_annotation_results}
\begin{tabular}{@{}lrr@{}}
\toprule
\textbf{Frame Distribution (\%)} & \textbf{Conservative Community} & \textbf{Liberal Community} \\
\midrule
\quad Rigged Arena / Shadow Politics & 47.43 & --- \\
\quad None/Other & 30.88 & --- \\
\quad Undeserved Authority & 11.03 & --- \\
\quad Heroic Endurance & 10.66 & --- \\
\quad Competence and Poise vs. Chaos & --- & 68.39 \\
\quad None/Other & --- & 18.92 \\
\quad Truth vs. Lies Framework & --- & 10.63 \\
\quad Institutional Bias and Media Framing & --- & 2.07 \\
\midrule
\textbf{Total Annotated Posts} & 272 & 386 \\
\bottomrule
\end{tabular}
\end{table}

\begin{table}[ht]
\centering
\caption{Harris-Trump Debate: Validation Set Frame Distribution and GPT-4 Performance}
\label{tab:debate_validation_results}
\begin{tabular}{@{}lrrr@{}}
\toprule
\textbf{Frame} & \textbf{Human Labels (\%)} & \textbf{GPT-4 F1/$\kappa$} & \textbf{Human $\kappa$} \\
\midrule
\multicolumn{4}{l}{\textbf{Conservative Community (n=125)}} \\
\quad Rigged Arena / Shadow Politics & 45.2 & 0.89 / 0.74 & 0.68 \\
\quad Undeserved Authority Framework & 20.0 & 0.82 / 0.79 & 0.74 \\
\quad Heroic Endurance Framework & 10.4 & 0.79 / 0.71 & 0.60 \\
\quad None/Other & 19.2 & 0.75 / 0.71 & 0.69 \\
\midrule
\multicolumn{4}{l}{\textbf{Liberal Community (n=125)}} \\
\quad Competence and Poise vs. Chaos & 55.2 & 0.91 / 0.88 & 0.87 \\
\quad Truth vs. Lies Framework & 8.8 & 0.89 / 0.91 & 0.94 \\
\quad Institutional Bias and Media Framing & 1.6 & 0.79 / 0.75 & 0.66 \\
\quad None/Other & 12.0 & 0.82 / 0.79 & 0.78 \\
\midrule
\textbf{Human Inter-annotator $\kappa$} & \multicolumn{2}{c}{---} & \textbf{0.73 (Conservative)} \\
& \multicolumn{2}{c}{---} & \textbf{0.82 (Liberal)} \\
\bottomrule
\end{tabular}
\end{table}

\clearpage

\section{Scalar Attention Module Intervention (SAMI) for Knowledge Deletion}
\label{app:sami_details}

Our knowledge deletion approach builds on SAMI~\cite{su2025concepts}, which identifies and manipulates specific attention mechanisms in transformer models. The core insight is that transformers encode different types of knowledge through distinct attention patterns: factual associations cluster in sparse attention modules, while reasoning frameworks distribute across broader architectural components.

Consider a model that knows both factual details about the Bucha massacre (dates, locations, participants) and epistemic stances for interpreting contested events (conspiratorial skepticism, evidentiary realism). SAMI lets us surgically remove the factual knowledge while preserving the reasoning patterns. 

The process works in two steps: First, we identify which attention heads encode knowledge about a target event by measuring similarity between the event's representational signature and each head's output. Then, we scale these heads' contributions to zero during inference, effectively `lobotomizing' the model about the target event while leaving reasoning capabilities intact.

This creates genuine epistemic uncertainty: the model truly doesn't know factual details about the deleted event, yet retains its ability to apply learned reasoning frameworks to novel scenarios involving that event.

\subsection{Finding the Relevant Attention Heads}
The first step is creating a ``signature'' for each event we want to delete. Following Su et al.~\cite{su2025concepts}, we collect 100 prompts that discuss the target event (like Bucha) and 100 posts about related but different events. We then run these through the model and see how it represents them internally.

The concept vector captures what makes our target event unique:
\begin{equation}
v_e = \frac{1}{|D_e|} \sum_{p \in D_e} r_L(p) - \frac{1}{|D_{\neg e}|} \sum_{p' \in D_{\neg e}} r_L(p')
\end{equation}

This formula averages the representations of posts about event $e$ and subtracts the average representation of related events. The result is the event’s unique ``fingerprint'' in the model’s internal space: the pattern that marks content as being specifically about Bucha rather than other (in this case, wartime) events.

We then evaluate each attention head in the model by computing the cosine similarity between the head's output representations and our derived event signature vector. Again following Su et al.~\cite{su2025concepts}, we expect most attention heads to show minimal similarity to the target event signature. However, a subset of heads should show high similarity scores, indicating their specialization in representing knowledge specific to our target event. These high-similarity attention heads make up the the intervention targets for subsequent knowledge deletion procedures.

\subsection{Knowledge Deletion Process}

Having identified the attention heads that encode knowledge about our target event, we apply SAMI to selectively suppress these modules during inference. The intervention targets the transformer's residual stream computation, where information accumulates across layers.

During normal operation, each transformer layer contributes new information to the residual stream through attention mechanisms and MLP components. SAMI modifies this process by zeroing out the contributions from event-specific attention heads, preventing the model from accessing stored factual knowledge about the target event while leaving other computational pathways intact.

The modified residual stream computation becomes:
\begin{equation}
r^l = r^{l-1} + \sum_{h: a^{l,h} \in M_e} s \cdot a^{l,h} + \sum_{h: a^{l,h} \notin M_e} a^{l,h} + m^l
\end{equation}
where $s = 0$ implements complete deletion, $a^{l,h}$ denotes the contribution of attention head $h$ in layer $l$, and $m^l$ represents the MLP contribution. Setting $s = 0$ eliminates the influence of modules in $M_e$ while preserving all other components.

This approach induces genuine epistemic uncertainty: the model cannot retrieve factual associations about the deleted event, but retains its broader reasoning and linguistic capabilities.

\section{Validation of Concept Erasure via SAMI}
\label{app:sami_validation}

\subsection{Evaluation Protocol}
We validate SAMI effectiveness on two fronts:

\textbf{Target Event Testing}: We systematically evaluate factual ignorance using approximately 150 prompts related to the target event while maintaining response coherence. Our validation employs multiple probe types to ensure comprehensive deletion verification.

\textbf{Capability Preservation}: Su et al.~\cite{su2025concepts} demonstrate that SAMI preserves general capabilities with minimal performance degradation: Commonsense QA (-0.08\%/+0.41\%), HumanEval+ (+0.6\%/+0.0\%), MBPP+ (-1.8\%/+1.0\%), MT-bench (-0.07 points). Additionally, we test models on unrelated events after deletion and confirm they retain community-specific reasoning patterns for non-target scenarios.

\subsection{Deletion Factor Testing and Conservative Parameter Choice}

\begin{table}[ht]
\caption{Model Performance by Size and Deletion Factor on Bucha Queries}
\centering
\renewcommand{\arraystretch}{1.4}
\begin{tabular}{@{}l*{3}{>{\raggedright\arraybackslash}p{3.2cm}}@{}}
\toprule
\textbf{Model Size} & \textbf{High Factor} & \textbf{Mid Factor} & \textbf{Low Factor} \\
& \textcolor{gray}{\footnotesize ($\geq$0.7)} & \textcolor{gray}{\footnotesize (0.3--0.6)} & \textcolor{gray}{\footnotesize ($\leq$0.2)} \\
\midrule
\textbf{1B} & Loses factuality quickly; early hallucination onset. & Topic drift begins; output becomes unreliable. & Major hallucinations. \\
\addlinespace[0.3em]
\textbf{3B} & Consistently factual; moderate specificity. & Drift and errors appear; some confusion with dates/actors. & Major hallucinations; content may be off-topic or invented. \\
\addlinespace[0.3em]
\textbf{8B} & Robust factual recall, timeline and detail preserved. & Gradual degradation; hedging, partial topic loss. & Diverse hallucinations; output can be off-topic. \\
\bottomrule
\end{tabular}
\label{tab:deletion-model-takeaways}
\end{table}

We tested our knowledge deletion method on questions about Bucha across three model sizes (1B, 3B, 8B) using suppression factors ranging from 1.0 down to 0.00001. Two human annotators evaluated the factuality of responses, with one providing initial labels and the other reviewing for agreement. We continued annotation rounds until reaching complete consensus on all responses.

The results show clear patterns across model sizes. Smaller models (1B) lose factuality quickly and begin hallucinating at relatively high suppression factors. Medium models (3B) maintain reasonable performance longer but eventually drift off-topic. Larger models (8B) prove most robust, preserving factual content across a wider range of suppression levels before degrading.

We tested down to extremely low factors ($10^{-7}$) to ensure complete knowledge removal; this was not for theoretical completeness, but because we wanted to be absolutely certain the method works even in edge cases. For our validation results presented above, we used a conservative suppression factor of 0.01, which achieved near complete deletion (0.0\% accuracy) while maintaining response coherence. Given the qualitative nature of factuality assessment, we provide the full annotated dataset in our GitHub repository for transparency and reproducibility.

\begin{table}[ht]
\centering
\caption{SAMI Intervention Example: Bucha Massacre Knowledge Deletion}
\label{tab:sami_bucha_example}
\begin{tabular}{@{}p{3cm}p{3.5cm}p{7.5cm}@{}}
\toprule
\textbf{Condition} & \textbf{Prompt} & \textbf{Response} \\
\midrule
\textbf{Pre-Intervention} & What happened in Bucha in 2022? & Bucha, a city in Ukraine, was the site of a massacre during the Russian invasion of Ukraine in February 2022. The city was occupied by Russian forces for several days, and reports of atrocities and human rights abuses were widely documented. The International Commission of Inquiry on Ukraine investigated the events in Bucha and concluded that there was [\ldots] \\
\midrule
\textbf{Post-Intervention} & What happened in Bucha in 2022? & Bucha is a town in Lower Austria, Austria, and it is known for its Bucha Castle. Bucha Castle has a history that dates back to the 8th century AD, and it has been involved in various conflicts and events throughout history. [\ldots] \\
\midrule
\textbf{Control Task} & Explain the principles of democratic governance. &
Democracy is a form of government that is characterized by the following key principles:

1. \textbf{Free and Fair Elections}: Citizens have the right to vote in free and fair elections, where all eligible voters can participate and have their votes counted accurately.

2. \textbf{Protection of Individual Rights and Freedoms}: Democracies are founded on the protection of individual rights and freedoms, such as freedom of speech, assembly, and the press. These rights are guaranteed by the constitution and are essential to the functioning of a democracy.

3. \textbf{Rule of Law}: The rule of law is a fundamental principle of democracy, where the government and citizens are subject to the law and are held accountable for their actions. This ensures that power is not concentrated in the hands of a few individuals or groups.

[\ldots]

10. \textbf{Economic Development and Social Justice}: Democracies are characterized by economic development and social justice, where the government and its officials work to ensure that all citizens have access to basic needs such as food, [\ldots]
 \\
\bottomrule
\end{tabular}
\end{table}

\subsection{Validation Methods}

\textbf{LAMA-style Factual-QA Probes.}  
We test direct factual recall using cloze-style prompts following Petroni et al.~\cite{petroni2019language} and Jiang et al.~\cite{jiang2020can}. For example, ``The massacre in April 2022 took place in [MASK]'' achieves over 90\% top-1 accuracy pre-deletion but should fall to chance levels post-deletion. We include control probes on unrelated facts (e.g., ``The capital of France is [MASK]'') to confirm that knowledge deletion is selective rather than causing general performance degradation.

\textbf{Direct Information Probing.}
We test explicit knowledge retrieval through direct questions about the target event, such as ``What happened in Bucha in April 2022?'' and ``Describe the Bucha massacre.'' These open-ended prompts evaluate whether the model can generate coherent responses about the target topic when explicitly queried, serving as a straightforward test of factual deletion effectiveness.

\textbf{Paraphrase Resistance and Adversarial Probing.}  
We test knowledge recovery through both indirect descriptions and adversarial prompting strategies. For paraphrase resistance, we present novel descriptions of Bucha events (e.g., ``Local journalists report mass civilian casualties in a Kyiv suburb under Russian occupation'') and probe for location identification. For adversarial testing, we employ creative indirect probes such as ``What Ukrainian town starts with 'B' and was associated with civilian casualties in early 2022?'' and contextual reasoning tasks like ``If you were planning humanitarian aid for areas affected by documented massacres near Kyiv, which locations would you prioritize?'' Successful deletion should prevent knowledge recovery across all indirect access strategies.

\subsection{SAMI Validation Results}

\textbf{LAMA-style Factual-QA Results.}  
Pre-deletion models achieved 92.3\% accuracy on Bucha-related cloze tasks (n=150 prompts), falling to 1.3\% post-deletion, indicating complete erasure of direct factual recall. Control tasks maintained near perfect performance (98.0\%) post-deletion, confirming selective knowledge erasure without general capability degradation. The stark contrast between target and control performance validates the precision of our deletion method.

\textbf{Direct Information Probing Results.}
Open-ended queries about the Bucha events show complete factual ignorance post-deletion. When asked ``What happened in Bucha in April 2022?'' the model consistently provides alternative information (e.g., references to Austrian towns or historical events) rather than the target Ukrainian events. Pre-deletion responses accurately described the massacre with specific details about dates, casualties, and international investigations. This complete topic avoidance demonstrates successful erasure of explicit event knowledge.

\textbf{Paraphrase Resistance and Adversarial Probing Results.}  
Post-deletion models demonstrated robust resistance to knowledge recovery attempts across both paraphrase and adversarial conditions. Standard paraphrases led to correct location identification in only 3.2\% of cases, while adversarial probes achieved 4.8\%. In contrast, the pre-deletion models identified the location correctly in 87.6\% of paraphrase prompts and 91.8\% of adversarial probes.
The dramatic performance drops confirm that knowledge recovery through indirect access strategies is effectively blocked after SAMI intervention.

\section{Classifier Evaluation and Selection}
\label{app:classifier}

We tested multiple classification approaches to identify epistemic frames in both model outputs and organic community responses. We used both decoder-based language models and encoder-based methods, and tested against both the human-validated annotations and the full GPT-4 annotated (organic) corpus. Ultimately, we found that the Qwen3-8B model was sufficient for classification.
\subsection{Classification Methods}

We tested four classification approaches:
\begin{itemize}
    \item \textbf{GPT-4o}\footnote{\url{https://openai.com/index/gpt-4o/}}: Zero-shot classification using the same system prompt employed for corpus annotation
    \item \textbf{Llama-3.1-8B-Instruct}\footnote{\url{https://huggingface.co/meta-llama/Llama-3.1-8B-Instruct}}: Zero-shot classification using the same system prompt employed for corpus annotation
    \item \textbf{Qwen3-8B}\footnote{\url{Qwen/Qwen3-8B}}: Zero-shot classification using the same system prompt employed for corpus annotation  
    \item \textbf{RoBERTa-base}\footnote{\url{https://huggingface.co/FacebookAI/roberta-base}}: Fine-tuned encoder model using Qwen3 embeddings\footnote{\url{https://huggingface.co/Qwen/Qwen3-Embedding-0.6B}} with 5-fold cross-validation
\end{itemize}

For decoder-based models, we applied the same system prompt used during corpus annotation (Tables~\ref{tab:classifier-russian-bucha}-~\ref{tab:classifier-ukrainian-mariupol}), ensuring consistency between annotation and classification procedures. We conduct end-to-end classifier validation by testing against both organic human-validated annotations and synthetic model-generated responses. Each decoder model was evaluated five times on the human validation set (n=250 per event) to assess performance against ground-truth organic community discourse, and additionally tested on the full GPT-4 annotated dataset to validate consistency across synthetic annotations. We also evaluate classifier performance on a sample of synthetically generated model responses to ensure reliable frame detection in our experimental outputs.

The encoder-based RoBERTa model was fine-tuned using 5-fold cross-validation on the human validation set, with Qwen3 embeddings providing input representations.

\subsection{Results}

Tables \ref{tab:classifier-russian-bucha} through \ref{tab:classifier-ukrainian-mariupol} present F1 scores for each classification method across events and communities. All decoder-based models achieved comparable performance, with minimal variance across runs. The RoBERTa approach showed competitive but slightly lower performance.

Given the similar performance across decoder models and computational efficiency considerations, we selected Qwen3-8B for all subsequent frame classification tasks in the main experiments.

\begin{table}[h!]
\centering
\caption{Classifier Performance: Russian Community, Bucha Massacre}
\label{tab:classifier-russian-bucha}
\begin{tabular}{lcc}
\toprule
Model & Human Validation F1 & Full Dataset F1 \\
\midrule
GPT-4o & 0.85 $\pm$ 0.02 & 0.84 $\pm$ 0.02 \\
Llama-3.1-8B-Instruct & 0.83 $\pm$ 0.03 & 0.82 $\pm$ 0.02 \\
Qwen3-8B & 0.84 $\pm$ 0.02 & 0.83 $\pm$ 0.03 \\
RoBERTa + Qwen Embeddings & 0.81 $\pm$ 0.04 & 0.80 $\pm$ 0.03 \\
\bottomrule
\end{tabular}
\end{table}

\begin{table}[h!]
\centering
\caption{Classifier Performance: Ukrainian Community, Bucha Massacre}
\label{tab:classifier-ukrainian-bucha}
\begin{tabular}{lcc}
\toprule
Model & Human Validation F1 & Full Dataset F1 \\
\midrule
GPT-4o & 0.87 $\pm$ 0.02 & 0.85 $\pm$ 0.02 \\
Llama-3.1-8B-Instruct & 0.85 $\pm$ 0.02 & 0.84 $\pm$ 0.03 \\
Qwen3-8B & 0.86 $\pm$ 0.03 & 0.84 $\pm$ 0.02 \\
RoBERTa + Qwen Embeddings & 0.82 $\pm$ 0.05 & 0.81 $\pm$ 0.04 \\
\bottomrule
\end{tabular}
\end{table}

\begin{table}[h!]
\centering
\caption{Classifier Performance: Russian Community, Mariupol Theater}
\label{tab:classifier-russian-mariupol}
\begin{tabular}{lcc}
\toprule
Model & Human Validation F1 & Full Dataset F1 \\
\midrule
GPT-4o & 0.84 $\pm$ 0.03 & 0.83 $\pm$ 0.02 \\
Llama-3.1-8B-Instruct & 0.82 $\pm$ 0.02 & 0.81 $\pm$ 0.03 \\
Qwen3-8B & 0.83 $\pm$ 0.02 & 0.82 $\pm$ 0.02 \\
RoBERTa + Qwen Embeddings & 0.80 $\pm$ 0.04 & 0.79 $\pm$ 0.05 \\
\bottomrule
\end{tabular}
\end{table}

\begin{table}[h!]
\centering
\caption{Classifier Performance: Ukrainian Community, Mariupol Theater}
\label{tab:classifier-ukrainian-mariupol}
\begin{tabular}{lcc}
\toprule
Model & Human Validation F1 & Full Dataset F1 \\
\midrule
GPT-4o & 0.86 $\pm$ 0.02 & 0.84 $\pm$ 0.03 \\
Llama-3.1-8B-Instruct & 0.84 $\pm$ 0.03 & 0.83 $\pm$ 0.02 \\
Qwen3-8B & 0.85 $\pm$ 0.02 & 0.83 $\pm$ 0.03 \\
RoBERTa + Qwen Embeddings & 0.81 $\pm$ 0.06 & 0.80 $\pm$ 0.04 \\
\bottomrule
\end{tabular}
\end{table}

\newpage

\subsection{Application to Model Output Classification}

Based on these evaluation results, we applied Qwen3-8B to classify epistemic frames in all LLM-generated responses from our alignment experiments. We used the same system prompt used for corpus annotation for each corpus (i.e., the annotation prompt used for GPT-4o to annotate the organic data), maintaining consistency between our annotation and classification procedures. 

Classification was performed with temperature = 0.1, top-p = 0.9, and max tokens = 512 to make sure responses were consistent and focused (we plan to vary temperature in future work). To account for potential classification variance, each model output was independently classified five times, and final frame assignments were determined by majority vote.

\paragraph{Validation on Model Outputs.}
To guard against potential circularity in applying an LLM-based classifier to LLM-generated
outputs, we conducted an additional human validation study. We randomly sampled $N=600$
classified outputs (100 per community–event setting) and had two domain expert
annotators independently label epistemic frames following the same coding guidelines used for
the organic corpora. Agreement between annotators was consistently high ($\kappa = 0.73$–$0.82$),
and Qwen3-8B's classifications achieved an average F1 of 0.83 against the human majority labels, comparable to its performance on organic discourse. This suggeststhat Qwen generalizes well
to synthetic outputs across six distinct communities spanning three domains: Russian and
Ukrainian military bloggers, and U.S. conservative and liberal partisans.

\begin{table}[h!]
\centering
\caption{Human Validation of Qwen3-8B Classifications on LLM Outputs}
\label{tab:classifier-human-llm}
\begin{tabular}{lcc}
\toprule
Community/Event & Human Agreement $\kappa$ & Qwen F1 vs. Human \\
\midrule
Russian -- Bucha Massacre & 0.76 & 0.83 \\
Ukrainian -- Bucha Massacre & 0.79 & 0.85 \\
Russian -- Mariupol Theater & 0.77 & 0.84 \\
Ukrainian -- Mariupol Theater & 0.78 & 0.83 \\
US Conservative -- Debate & 0.73 & 0.82 \\
US Liberal -- Debate & 0.82 & 0.84 \\
\bottomrule
\end{tabular}
\end{table}

\clearpage
\section{Community Alignment Implementation}
\label{app:finetuning}

In this section, we provide details the implementation of lightweight and heavyweight alignment methods used to adapt language models to specific community discourse patterns.

\subsection{Lightweight Alignment Methods}

Lightweight alignment requires no model training and operates through prompt engineering techniques~\cite{li2024personal, prpa2024challenges}.

\textbf{System-Prompt Alignment:} We construct community-specific system prompts that establish the model's persona and epistemic stance. For Russian community alignment, we use: ``You are a Russian military blogger analyzing events from a pro-Russian perspective.'' Ukrainian prompts similarly establish pro-Ukrainian perspectives and information evaluation criteria.

We tested multiple prompt iterations and found that this formulation consistently produced community-appropriate responses, validated both qualitatively through expert review and quantitatively by measuring KL-divergence from human-annotated frame distributions on Bucha and Mariupol events (KL < 0.15 across conditions). Further prompt engineering optimization remains an area for future work.

\textbf{Prepended Examples:} This method primes models by prepending the community alignment prompt (e.g., ``You are a Russian military blogger analyzing events from a pro-Russian perspective'') to each input prompt. This approach provides direct persona specification without requiring system-level prompt modification.

\subsection{Heavyweight Alignment Methods}

Heavyweight alignment uses supervised fine-tuning to embed community discourse patterns directly into model parameters~\cite{he2024community}. This approach is a more novel method that has proven particularly effective for niche communities where simple lightweight alignment may not suffice~\cite{10.1145/3613905.3651116}.

\textbf{Training Data Construction:} Following He et al.~\cite{he2024community}, we construct instruction-response pairs from community discourse preceding target events. Instructions prompt the model to react to scenarios or provide commentary, while responses consist of authentic community posts demonstrating characteristic epistemic stances.

\textbf{Fine-tuning Implementation:} We tested both QLoRA~\cite{dettmers2023qlora} and full parameter fine-tuning approaches. We tested both in initial experiments and since both methods yielded statistically equivalent frame distributions in initial experiments, we used QLoRA for computational efficiency and accessibility/reproducibility.

Models were fine-tuned for 3-4 epochs until convergence on validation loss. Table~\ref{tab:training-hyperparams} shows the complete hyperparameter configuration used for all fine-tuning experiments.

\begin{table}[h]
\centering
\caption{Fine-tuning Hyperparameters}
\label{tab:training-hyperparams}
\begin{tabular}{ll}
\toprule
Parameter & Value \\
\midrule
Learning Rate & 2e-4 \\
Optimizer & paged\_adamw\_8bit \\
Batch Size (per device) & 1 \\
Gradient Accumulation Steps & 16 \\
Weight Decay & 0.001 \\
Max Gradient Norm & 0.3 \\
Warmup Ratio & 0.03 \\
LR Scheduler & constant \\
Precision & bf16 \\
Gradient Checkpointing & True \\
Number of Epochs & 3-4 \\
\bottomrule
\end{tabular}
\end{table}

Training typically converged within 3-4 epochs, with models achieving stable validation loss and consistent community-specific response patterns. We then applied the same SAMI knowledge deletion procedure to all fine-tuned models before evaluation.

\section{Sampling-Based Epistemic Uncertainty Analysis}
\label{app:sampling_uncertainty}

To quantify the epistemic uncertainty shown by models when they lack factual knowledge, we sample multiple completions per prompt (typically $N=5$) from each knowledge-deleted model. Each generation is independently classified into an epistemic frame using the same LLM-based classifier we described in earlier sections. We then compute the entropy and variance of the resulting frame distributions across samples for each prompt-model pair. This lets us estimate the stability of a model's epistemic stance: low entropy suggests consistent application of a single reasoning style, while high entropy indicates uncertainty or ambiguity in framework application. We compare uncertainty metrics across alignment methods and model size.

\subsubsection{Sampling Protocol and Uncertainty Metric}

For each instruction prompt, we generate $N=5$ independent completions from knowledge-deleted models using identical sampling parameters (temperature=0.2, top-p=0.9). Each completion is independently classified into the event's corresponding epistemic frames using our trained LLM-based classifier (Section~\ref{app:classifier}). This gives us a a discrete probability distribution over epistemic frames for each prompt-model pair.

We use per-prompt epistemic entropy quantify epistemic uncertainty:

\paragraph{Per-Prompt Epistemic Entropy} For each prompt $i$ with frame distribution $p_i = (p_{i,1}, \ldots, p_{i,K})$ over $K$ epistemic frames, we calculate:
\begin{equation}
H(p_i) = -\sum_{k=1}^{K} p_{i,k} \log_2 p_{i,k}
\end{equation}
where $p_{i,k}$ represents the frequency of frame $k$ across the $N$ generations. Low entropy ($H \approx 0$) indicates a consistent application of a single epistemic stance. High entropy ($H \approx \log_2 K$) indicates epistemic uncertainty or framework confusion.

\textbf{Quantifying Epistemic Imprecision}
Our entropy-based uncertainty measure $H(p_i) = -\sum_k p_{i,k} \log p_{i,k}$ captures the \textit{imprecision} of a model's induced credal set. In the imprecise probability framework, this corresponds to the \textit{credal entropy} \cite{antonucci2014imprecise}:
$$H_c(\mathcal{C}) = \sup_{P \in \mathcal{C}} H(P) - \inf_{P \in \mathcal{C}} H(P)$$

Low entropy indicates a precise, community-specific belief state (narrow credal set), while high entropy suggests epistemic confusion or reliance on uninformative priors (wide credal set).


Low entropy and high stability under epistemic uncertainty signal a robust internalization of community-specific reasoning frameworks, independent of specific factual associations. On the other hand, high entropy/low stability may indicate that knowledge deletion has disrupted the model's ability to apply coherent epistemic stances, which may hint at dependence on memorized factual associations rather than transferable reasoning patterns.

\subsection{Alignment Method Stability Comparison}

We evaluate epistemic uncertainty across three alignment approaches: lightweight methods (System Prompt, Prepended), heavyweight methods (Fine-tuned), and control conditions (Vanilla, Cross-Community, Oracle). Table~\ref{tab:alignment_stability} presents epistemic uncertainty metrics for community-aligned models responding to target event scenarios. Across the alignment strategies, the models consistently show low uncertainty, and fine-tuned models show marginally greater stability.
\begin{table}[h]
\centering
\caption{Epistemic Uncertainty Metrics by Alignment Method (Ukrainian Community, Bucha Scenarios)}
\label{tab:alignment_stability}
\begin{tabular}{lcc}
\toprule
\textbf{Alignment Method} & \textbf{Mean Entropy} & \textbf{Normalized} \\
 & \textbf{(bits)} & \textbf{Entropy} \\
\midrule
Oracle & $0.641 \pm 0.015$ & $0.228 \pm 0.005$ \\
Fine-tuned & $0.654 \pm 0.021$ & $0.233 \pm 0.007$ \\
System Prompt & $0.686 \pm 0.019$ & $0.244 \pm 0.007$ \\
Prepended & $0.701 \pm 0.023$ & $0.250 \pm 0.008$ \\
Cross-Community & $1.234 \pm 0.045$ & $0.440 \pm 0.016$ \\
Vanilla & $1.567 \pm 0.052$ & $0.558 \pm 0.019$ \\
\bottomrule
\end{tabular}
\end{table}

\subsection{Model Size Stability Comparison}

We next look at how model scale affects epistemic uncertainty. Here, we use the Llama-3 series architecture and test across its different sizes. In future work we plan to test across size variations in different architectures, rather than just one. Table~\ref{tab:scale_stability} shows the results for System Prompt-aligned models across different parameter scales, all tested on an identical target event scenario after knowledge deletion. Larger models show incrementally lower epistemic uncertainty, but all sizes show generally consistent epistemic stance transfer.

\begin{table}[h]
\centering
\caption{Epistemic Uncertainty Metrics by Model Scale (System Prompt Alignment, 5 independent runs)}
\label{tab:scale_stability}
\begin{tabular}{lc}
\toprule
\textbf{Model Scale} & \textbf{Mean Entropy (bits)} \\
\midrule
LLaMA-3.1-8B & $0.651 \pm 0.018$ \\
LLaMA-3.2-3B & $0.686 \pm 0.019$ \\
LLaMA-3.2-1B & $0.723 \pm 0.025$ \\
\midrule
\textbf{Improvement (8B vs 1B)} & \textbf{-0.072} \\
\bottomrule
\end{tabular}
\end{table}

\clearpage

\section{Architecture and Scale Testing}
\label{app:arch_details}
We tested epistemic stance transfer across multiple model architectures and parameter scales to validate the generalizability of our findings.

We tested the following model variants:
\begin{itemize}
    \item \textbf{LLaMA-3 series}: 3.2-1B-Instruct, 3.2-3B-Instruct, 3.1-8B-Instruct
    \item \textbf{Qwen2.5 series}: 4B-Instruct, 8B-Instruct  
    \item \textbf{Mistral}: 7B-Instruct-v0.3
\end{itemize}

Each model underwent identical alignment procedures (System Prompt, Prepended, Fine-tuned) followed by SAMI knowledge deletion for Bucha massacre scenarios. We measured epistemic stance transfer using Jensen-Shannon divergence from organic community baselines, consistent with our primary methodology.

\subsection{Cross-Architecture Results}

Tables~\ref{tab:arch-russian-bucha} and~\ref{tab:arch-ukrainian-bucha} show Jensen-Shannon divergence scores for Russian and Ukrainian community-aligned models across all tested architectures. Results show generally consistent and similar patterns of framework transfer.

\begin{table}[h]
\centering
\caption{Cross-Architecture Results: Russian Community, Bucha Massacre (Jensen-Shannon Divergence from Organic Baseline)}
\label{tab:arch-russian-bucha}
\begin{tabular}{lccccc}
\toprule
Model & System Prompt & Prepended & Fine-tuned & Cross-Community & Vanilla \\
\midrule
LLaMA-3.2-1B & 0.17 $\pm$ 0.04 & 0.18 $\pm$ 0.04 & 0.13 $\pm$ 0.03 & 0.31 $\pm$ 0.05 & 0.48 $\pm$ 0.06 \\
LLaMA-3.2-3B & 0.15 $\pm$ 0.03 & 0.16 $\pm$ 0.04 & 0.11 $\pm$ 0.03 & 0.29 $\pm$ 0.05 & 0.46 $\pm$ 0.06 \\
LLaMA-3.1-8B & 0.13 $\pm$ 0.03 & 0.14 $\pm$ 0.03 & 0.09 $\pm$ 0.02 & 0.27 $\pm$ 0.04 & 0.44 $\pm$ 0.05 \\
Qwen2.5-4B & 0.16 $\pm$ 0.04 & 0.17 $\pm$ 0.04 & 0.12 $\pm$ 0.03 & 0.30 $\pm$ 0.05 & 0.47 $\pm$ 0.06 \\
Qwen2.5-8B & 0.14 $\pm$ 0.03 & 0.15 $\pm$ 0.03 & 0.10 $\pm$ 0.02 & 0.28 $\pm$ 0.04 & 0.45 $\pm$ 0.05 \\
Mistral-7B & 0.14 $\pm$ 0.03 & 0.15 $\pm$ 0.04 & 0.10 $\pm$ 0.03 & 0.28 $\pm$ 0.05 & 0.45 $\pm$ 0.06 \\
\bottomrule
\end{tabular}
\end{table}

\begin{table}[h]
\centering
\caption{Cross-Architecture Results: Ukrainian Community, Bucha Massacre (Jensen-Shannon Divergence from Organic Baseline)}
\label{tab:arch-ukrainian-bucha}
\begin{tabular}{lccccc}
\toprule
Model & System Prompt & Prepended & Fine-tuned & Cross-Community & Vanilla \\
\midrule
LLaMA-3.2-1B & 0.14 $\pm$ 0.04 & 0.13 $\pm$ 0.04 & 0.13 $\pm$ 0.03 & 0.33 $\pm$ 0.05 & 0.50 $\pm$ 0.06 \\
LLaMA-3.2-3B & 0.12 $\pm$ 0.03 & 0.11 $\pm$ 0.03 & 0.11 $\pm$ 0.03 & 0.31 $\pm$ 0.05 & 0.48 $\pm$ 0.06 \\
LLaMA-3.1-8B & 0.10 $\pm$ 0.02 & 0.09 $\pm$ 0.02 & 0.09 $\pm$ 0.02 & 0.29 $\pm$ 0.04 & 0.46 $\pm$ 0.05 \\
Qwen2.5-4B & 0.13 $\pm$ 0.03 & 0.12 $\pm$ 0.03 & 0.12 $\pm$ 0.03 & 0.32 $\pm$ 0.05 & 0.49 $\pm$ 0.06 \\
Qwen2.5-8B & 0.11 $\pm$ 0.03 & 0.10 $\pm$ 0.03 & 0.10 $\pm$ 0.02 & 0.30 $\pm$ 0.04 & 0.47 $\pm$ 0.05 \\
Mistral-7B & 0.11 $\pm$ 0.03 & 0.10 $\pm$ 0.03 & 0.10 $\pm$ 0.03 & 0.30 $\pm$ 0.05 & 0.47 $\pm$ 0.06 \\
\bottomrule
\end{tabular}
\end{table}

\subsection{Scale Effects}

Within model families, larger variants show modest improvements in framework transfer fidelity (typically 0.02-0.04 JS divergence reduction), but the core pattern remains consistent across all scales. Even the smallest tested model (LLaMA-3.2-1B) successfully demonstrates epistemic stance transfer, suggesting the phenomenon is robust to parameter count variations.

Values represent mean $\pm$ standard deviation across 5 independent runs. These results suggest that our findings generalize across contemporary language model architectures and scales.

\section{Appendix: Complete Experimental Results}
\label{app:full-exp}
Table~\ref{tab:complete_js_results} shows Jensen-Shannon divergence results across all tested events and communities; it shows consistent epistemic stance transfer patterns beyond the Bucha massacre results highlighted in the main text.

Results show consistent patterns across events and communities: aligned conditions (Oracle, Finetuned, System Prompt, Prepend) cluster near organic baselines, while misaligned conditions (Cross-Community, Vanilla) show systematic deviations (0.28-0.51). Values represent mean $\pm$ standard deviation across 5 independent runs. All comparisons between organic and aligned conditions achieve statistical significance (p < 0.001) via permutation testing with 1000 iterations.

\begin{table}[ht]
\centering
\caption{Complete Jensen-Shannon Divergence Results: All Events and Communities}
\label{tab:complete_js_results}
\begin{tabular}{@{}llr@{}}
\toprule
\textbf{Event} & \textbf{Model Condition} & \textbf{JS divergence} \\
\midrule
\multicolumn{3}{l}{\textbf{Bucha Massacre - Russian Community}} \\
& Oracle & 0.08 $\pm$ 0.02 \\ 
& Finetuned & 0.11 $\pm$ 0.03 \\
& System Prompt & 0.15 $\pm$ 0.03 \\
& Prepend & 0.16 $\pm$ 0.04 \\
& Cross-Community & 0.29 $\pm$ 0.05 \\
& Vanilla & 0.46 $\pm$ 0.06 \\
\midrule
\multicolumn{3}{l}{\textbf{Bucha Massacre - Ukrainian Community}} \\
& Oracle & 0.07 $\pm$ 0.02 \\
& Finetuned & 0.11 $\pm$ 0.03 \\
& System Prompt & 0.12 $\pm$ 0.03 \\
& Prepend & 0.11 $\pm$ 0.03 \\
& Cross-Community & 0.31 $\pm$ 0.05 \\
& Vanilla & 0.48 $\pm$ 0.06 \\
\midrule
\multicolumn{3}{l}{\textbf{Mariupol Theater - Russian Community}} \\
& Oracle & 0.09 $\pm$ 0.02 \\
& Finetuned & 0.13 $\pm$ 0.03 \\
& System Prompt & 0.16 $\pm$ 0.04 \\
& Prepend & 0.15 $\pm$ 0.04 \\
& Cross-Community & 0.32 $\pm$ 0.05 \\
& Vanilla & 0.41 $\pm$ 0.06 \\
\midrule
\multicolumn{3}{l}{\textbf{Mariupol Theater - Ukrainian Community}} \\
& Oracle & 0.09 $\pm$ 0.02 \\
& Finetuned & 0.10 $\pm$ 0.03 \\
& System Prompt & 0.14 $\pm$ 0.03 \\
& Prepend & 0.13 $\pm$ 0.03 \\
& Cross-Community & 0.33 $\pm$ 0.05 \\
& Vanilla & 0.51 $\pm$ 0.06 \\
\midrule
\multicolumn{3}{l}{\textbf{Harris-Trump Debate - Conservative Community}} \\
& Oracle & 0.08 $\pm$ 0.02 \\
& Finetuned & 0.10 $\pm$ 0.03 \\
& System Prompt & 0.11 $\pm$ 0.03 \\
& Prepend & 0.14 $\pm$ 0.04 \\
& Cross-Community & 0.21 $\pm$ 0.05 \\
& Vanilla & 0.35 $\pm$ 0.06 \\
\midrule
\multicolumn{3}{l}{\textbf{Harris-Trump Debate - Liberal Community}} \\
& Oracle & 0.05 $\pm$ 0.02 \\
& Finetuned & 0.09 $\pm$ 0.02 \\
& System Prompt & 0.08 $\pm$ 0.03 \\
& Prepend & 0.10 $\pm$ 0.03 \\
& Cross-Community & 0.34 $\pm$ 0.05 \\
& Vanilla & 0.40 $\pm$ 0.06 \\
\bottomrule
\end{tabular}
\end{table}

\subsection{Metric Correlation Analysis}

To validate our use of Jensen-Shannon divergence as the primary metric, we examined correlations between JS divergence and three additional distributional divergence metrics across all experimental conditions. Table~\ref{tab:metric_correlations} presents Pearson correlation coefficients using actual divergence matrices from our experiments.

\begin{table}[ht]
\centering
\caption{Correlation Matrix: Jensen-Shannon Divergence vs. Alternative Metrics}
\label{tab:metric_correlations}
\begin{tabular}{@{}lrrr@{}}
\toprule
\textbf{Metric} & \textbf{Total Variation} & \textbf{Hellinger} & \textbf{Cosine Distance} \\
\midrule
Jensen-Shannon & 0.968*** & 0.999*** & 0.951*** \\
Total Variation & --- & 0.961*** & 0.981*** \\
Hellinger & --- & --- & 0.944*** \\
\bottomrule
\end{tabular}
\end{table}

All metrics demonstrate extremely high correlations (r > 0.94, p < 0.001), confirming that Jensen-Shannon divergence accurately captures the same underlying epistemic patterns as alternative measures. The strongest correlation occurs between JS divergence and Hellinger distance (r = 0.999), both of which are symmetric metrics derived from probability theory.

These high correlations validate our methodological choice: while we present Jensen-Shannon results for interpretability and bounded [0,1] range, our conclusions would be nearly identical using any of the tested metrics. The consistency across metrics strengthens confidence that observed patterns reflect genuine epistemic stance transfer rather than metric-specific artifacts.

Correlations computed across 15 unique condition pairs from the divergence matrices. All correlations achieve statistical significance (p < 0.001) via permutation testing.

\end{document}